\documentclass[onefignum,onetabnum]{siamonline250211}

\usepackage{lipsum}
\usepackage{amsfonts}
\usepackage{graphicx}
\usepackage{epstopdf}
\usepackage{algorithmic}
\usepackage[caption=false]{subfig}
\usepackage{overpic}

\ifpdf
  \DeclareGraphicsExtensions{.eps,.pdf,.png,.jpg}
\else
  \DeclareGraphicsExtensions{.eps}
\fi

\usepackage{enumitem}
\setlist[enumerate]{leftmargin=.5in}
\setlist[itemize]{leftmargin=.5in}

\newsiamremark{remark}{Remark}
\newsiamremark{hypothesis}{Hypothesis}
\crefname{hypothesis}{Hypothesis}{Hypotheses}
\newsiamthm{claim}{Claim}
\newsiamremark{fact}{Fact}
\crefname{fact}{Fact}{Facts}

\headers{A Morse-Bott Framework for Blind Inverse Problems}{M. H. Nguyen, E. Pauwels and P. Weiss}

\title{A Morse-Bott Framework for Blind Inverse Problems \\ Local Recovery Guarantees and the Failure of the MAP \thanks{Submitted to the editors DATE.
\funding{The authors acknowledge a support from the ANR Micro-Blind (ANR-21-CE48-0008) and from the ANR CLEAR-Microscopy (ANR-25-CE45-3780). 
The computational resources were provided by the IDRIS-CNRS under the allocation 2021-AD011012210R3.}}}

\author{Minh-Hai Nguyen \thanks{IRIT, Toulouse University, Toulouse, France
  (\email{minh-hai.nguyen@irit.fr}, \email{pierre.armand.weiss@gmail.com}).}
\and Edouard Pauwels \thanks{Toulouse School of Economics, Université Toulouse Capitole et IUF, Toulouse, France 
  (\email{edouard2.pauwels@ut-capitole.fr}).}
\and Pierre Weiss \footnotemark[2]}

\usepackage{amsopn}

\newcommand{\eqdef}{\stackrel{\mathrm{def}}{=}}

\newcommand{\Normal}[1]{\mathcal{N}\left( {#1} \right)}

\newcommand{\Id}{\mathrm{Id}}

\newcommand{\norm}[1]{\left\Vert #1 \right\Vert}

\newcommand{\x}{\boldsymbol{x}}
\newcommand{\h}{\boldsymbol{h}}
\newcommand{\y}{\boldsymbol{y}}

\newcommand{\w}{\boldsymbol{w}}

\newcommand{\btheta}{\boldsymbol{\theta}}

\newcommand{\Lc}{L}%{\mathcal{L}}

\newcommand{\xmap}{\hat{x}_{{\scriptscriptstyle\mathrm{MAP}}}}

\newcommand{\hmap}{\hat{h}_{\scriptscriptstyle{\mathrm{MAP}}}}

\newcommand{\conv}{\star}

\newcommand{\R}{\mathbb{R}}

\DeclareMathOperator*{\argmin}{arg\,min}
\DeclareMathOperator*{\prox}{prox}

\newcommand{\etal}{\textit{et al.}}

\newcommand{\eg}{\textit{e.g.}, }

\usepackage{multirow}
\usepackage{tabularx}
\ifpdf
\hypersetup{
    pdftitle={A Morse--Bott Framework for Blind Inverse Problems: Local Recovery Guarantees and Pitfalls in Deconvolution},
    pdfauthor={M. H. Nguyen, E. Pauwels and P. Weiss}
}
\fi

\begin{document}

\maketitle

% REQUIRED
\begin{abstract}
    Maximum A Posteriori (MAP) estimation is a cornerstone framework for blind inverse problems, where an image and a forward operator are jointly
    estimated as the maximizers of a posterior distribution. In applications such as blind deblurring, this principle is used to recover sharp images
    from degraded observations. In this paper, we analyze the recovery guarantees of MAP-based methods by adopting a \emph{Morse--Bott framework}. We
    model the image potential as a Morse--Bott function, where natural images are modeled as residing locally on a critical submanifold. This means
    that while the potential is locally flat along the ``natural'' directions of the image manifold, it is strictly convex in the directions normal to it.
    We demonstrate that this Morse--Bott hypothesis aligns with the structural properties of state-of-the-art learned priors, a finding we validate
    through an experimental analysis of the potential landscape and its Hessian spectrum.
    Our theoretical results show that, in a neighborhood of the ground-truth image and operator, the posterior admits local minimizers that are
    stable both with respect to initialization (gradient descents converge to the same minimizer) and to small perturbations of the data (solutions
    vary smoothly with the observations). This local stability potentially provides a theoretical justification for the empirical success of well designed
    gradient-based optimization in these settings.
    However, we also demonstrate that this local stability is a \textbf{local} property: the ``blurry trap'', well-known for sparse priors in blind
    deconvolution, persists even with state-of-the-art learned priors.
    Our findings demonstrate that the failure of MAP in blind deconvolution is not a limitation of prior quality, but an intrinsic characteristic of
    the landscape.
    We conclude that successful recovery depends on strategic initialization around favorable local minima. We describe an heuristic strategy which is validated with
    numerical experiments on both synthetic and real-world data.
    The code for reproducing the numerical experiments in this paper is available at \url{https://github.com/mh-nguyen712/blind_deblur_diffusion}
\end{abstract}

% REQUIRED

% REQUIRED

% ------------------------------------------------------------
% ------------------------------------------------------------
% ------------------------------------------------------------
% --------- INTRODUCTION ---------------
% ------------------------------------------------------------
% ------------------------------------------------------------
% ------------------------------------------------------------
\section{Introduction}
\label{sec:intro}

Consider a parameterized linear forward operator \(A(\theta):\R^N\to \R^M\), where \(\theta \in \R^P\) is a parameter vector describing the action of the
operator. Blind inverse problems in imaging consist in recovering an image \(\bar x \in \R^N\) and the parameter \(\bar \theta\) from an indirect
observation \(y\) given by
\begin{equation}\label{eq:forward_model}
    y = A(\bar \theta)(\bar x) + b
\end{equation}
where \(b\) represents additive noise. This problem is fundamental in areas such as biomedical imaging, astronomy, and computer vision. The parameter
\(\bar \theta\) typically describes operator properties such as point spread functions in optics, unknown projection angles in tomography, or
sensitivity maps in MRI.

It is well-known that blind inverse problems are severely ill-posed, as there exist infinitely many pairs \((x, \theta)\) capable of explaining the
observation \(y\). Consequently, algorithms accompanied by theoretical recovery guarantees for \((\bar x, \bar \theta)\) are scarce and often rely on
restrictive or unrealistic assumptions~\cite{kech2017optimal,krahmer2021convex,debarnot2023blind,ling2018self}.

Taking a Bayesian viewpoint, we assume that \(\bar x\) and \(\bar \theta\) are realizations of two \emph{independent} random vectors \(\x\) and \(\btheta\)
with prior distributions \(p_{\x}\) and \(p_{\btheta}\). Using Bayes' formula, the posterior distribution is:
\begin{equation}
    p(x, \theta \vert y) \propto p(y \vert x, \theta) p_{\x}(x) p_{\btheta}(\theta).
\end{equation}
Under the assumption of white Gaussian noise \(b\sim \Normal{0, \sigma^2 \Id}\), the negative-log-posterior reads:
\begin{equation}
    \label{eq:nlp}
    \Lc(x, \theta;y) = \frac{\norm{A(\theta)(x) - y}_2^2}{2\sigma^2}  - \log p_{\x}(x) - \log p_{\btheta}(\theta).
\end{equation}
In all the paper, we assume that the parameter \(\btheta\) is uniformly distributed over a compact set \(\Theta \subset \R^P\), leading to a constant
potential \(-\log p_{\btheta}(\theta)\) that can be ignored in \cref{eq:nlp}. The negative-log-prior \(-\log p_{\x}(x)\) is called potential.

The Maximum A Posteriori (MAP) estimator seeks to minimize \cref{eq:nlp} with respect to the pair \((x, \theta)\). While handcrafted priors --
promoting sparsity in spatial~\cite{pan2016blind}, wavelet~\cite{pustelnik2016wavelet}, or gradient domains~\cite{chan1998total} -- have historically
dominated the field, recent advances in generative modeling~\cite{song2021scorebased,rombach2022high} have demonstrated unprecedented capability in
capturing the complexity of natural images.
Despite the power of these learned priors, their interaction with the MAP framework remains poorly understood. Specifically, the field lacks a
rigorous explanation for why global MAP optimization leads to degenerate solutions, while local optimization -- when properly initialized -- remains
remarkably effective.

Although the theoretical results derived hereafter extend to general blind inverse problems, we will focus on blind deconvolution in the experimental part.
For this specific problem, popular approaches for describing the kernel prior include a direct modeling in the pixel domain with non-negativity and
normalization constraints (simplex)~\cite{chan1998total,pan2014deblurring}, more specific parametric models such as linear motion and out-of-focus
blurs~\cite{oliveira2013parametric}, diffraction-limited families~\cite{shajkofci2020spatially,debarnot2022deep} or recent generative
modeling~\cite{blindDPS}. In this setting, it has often been observed that global MAP optimization leads to degenerate
solutions~\cite{levin2009understanding,benichoux2013fundamental}, while local optimization with careful initialization may succeed.

\paragraph{Contributions}
The progress in learned priors raises a fundamental question: \emph{How does the geometry of learned prior landscapes shape the posterior in blind
inverse problems, and can we formalize the conditions for local recovery versus global failure?} This paper addresses this above question through a
rigorous theoretical and empirical analysis of the posterior landscape. Our contributions are as follows:
\begin{itemize}
    \item We introduce a \emph{Morse--Bott modeling hypothesis} of image potentials. We formalize the intuition that images lie on
        low-dimensional manifolds by modeling the potential as a Morse--Bott function, where the ``natural image manifold'' corresponds to a
        non-degenerate critical set.
    \item We provide \emph{empirical evidence} showing that this hypothesis fits state-of-the-art diffusion priors, utilizing a novel analysis of the
        Hessian spectral gap to verify the non-degeneracy conditions required by the theory.
    \item We establish \emph{local recovery guarantees} for general inverse and blind inverse problems.
        Under a Morse--Bott assumption on the potentials, we prove that the ground-truth pair \((\bar x, \bar \theta)\), with \(\bar x\) a
        \emph{second-order critical point} of the potential, acts as a stable local minimizer of the joint posterior, provided the forward operator
        satisfies certain injectivity conditions quantified through explicit stability constants.
        The stability notion is twofold: (i) with respect to initialization, in the sense that gradient descent starting in a neighborhood of \((\bar
        x, \bar \theta)\) converges to the same minimizer, and (ii) with respect to small perturbations of the measurements \(y\), in the sense that the
        recovered solution depends smoothly on the data.
    \item We provide numerical validation of these conditions for various blur kernels in the context of blind deconvolution.
    \item In the specific case of blind deconvolution, we identify the \emph{blurry trap} as a fundamental limitation of global MAP.
        We validate empirically that for natural image distributions, learned potentials consistently favor blurry images over sharp ones.
        This explains why the MAP coincides with a ``no-blur'' solution regardless of the potential's generative quality.
	\item We propose an \emph{informed initialization strategy} to navigate this landscape. We describe a heuristic inspired by our posterior landscape analysis. 
		Empirical experiments demonstrate that successful recovery in blind deconvolution can be achieved with the proposed basin selection strategy. These experiments effectively bridges our theoretical conditions with a
        practical ``large-to-small'' kernel initialization.
\end{itemize}

\section{The Morse--Bott Landscape}
To analyze the convergence properties of blind deconvolution, we must formalize the geometry of the potential \(q(x) = -\log p_{\x}(x)\). Strong
recovery guarantees for linear inverse problems have traditionally relied on convex, non-smooth potentials---a framework instrumental to the field of
compressed sensing~\cite{candes2005decoding, vaiter2017model}. However, these handcrafted potentials are increasingly recognized as insufficient for
capturing the rich, high-dimensional complexity of natural images.

In this work, we pivot from convex to non-convex potentials.
We adopt the \emph{Morse--Bott property} as a modeling hypothesis to characterize the local geometry of modern learned potentials.
This framework accounts for the fact that natural images do not occupy the entire ambient space, but rather reside on a low-dimensional
manifold~\cite{popeintrinsic}.

\begin{definition}[Morse--Bott Potential]\label{def:morse_bott}
    Let \(q: \mathbb{R}^N \to \mathbb{R}\) be a \(\mathcal{C}^2\) potential. We denote the set of its critical points as \(\mathcal{C} = \{\bar x \in
    \mathbb{R}^N \mid \nabla q(\bar x) = 0\}\).
    The potential \(q\) is said to be \emph{Morse--Bott} locally at \(\bar x \in \mathcal{C}\) if:
    \begin{enumerate}
        \item There exists a smooth submanifold \(\mathcal{M}\) such that \(\mathcal{C} = \mathcal{M}\) locally around \(\bar x\).
        \item The Hessian \(\nabla^2 q(\bar x)\) is non-degenerate in the directions normal to \(\mathcal{M}\). Equivalently, the kernel of the Hessian
            coincides exactly with the tangent space, \(T_{\bar x} \mathcal{M}\), to \(\mathcal{M}\) at \(\bar x\):
            \[ \ker(\nabla^2 q(\bar x)) = T_{\bar x} \mathcal{M}. \]
    \end{enumerate}
    The potential \(q\) is called Morse--Bott, if it is Morse--Bott at any \(\bar{x} \in \mathcal{C}\).
\end{definition}
Examples of Morse--Bott potentials include functions of the form \(q(x) = \norm{Wx}^2\) where \(W\) is a rank-deficient matrix. More generally, given a
smooth submanifold \(\mathcal{M}\), the distance function to \(\mathcal{M}\), \(d_{\mathcal{M}}\) is Morse--Bott locally at each \(\bar x \in \mathcal{M}\)
\cite[Problem 6-5]{lee2003smooth}. Indeed, \(q\) being locally Morse--Bott at a local minimizer \(\bar{x} \in \mathcal{M}\) expresses roughly the fact
that local minimizers of \(q\) form smooth submanifold locally around \(\bar{x}\), such that \(x \mapsto q(x) - q(\bar{x})\) is of order
\(d^2_{\mathcal{M}}(x)\) up to positive constants~\cite{rebjock2024fast}.

Thus, the Morse--Bott framework serves as a surrogate model to express the compatibility between the geometry of the image manifold and the
transverse curvature of the potential.
In an imaging context, this agrees with the well established fact that natural images live on low-dimensional structures~\cite{popeintrinsic}.
In the following sections, we explore how state-of-the-art diffusion-based potentials align with this modeling hypothesis and provide
additional insights on the potential's landscape.

\subsection{Learned Potentials exhibit the local Morse--Bott property}
To provide empirical support for the Morse--Bott modeling choice, we conduct an empirical analysis of the potential landscape \(q\).
We utilize state-of-the-art pretrained diffusion models from~\cite{song2021scorebased,Karras2022edm} on the FFHQ, ImageNet, and AFHQ datasets at
various resolutions, see~\cref{sec:experimental_setup} for more details.
For simplicity, we refer to the potential \(q\) associated with these models as FFHQ-256, FFHQ-64, ImageNet-64, and AFHQ-64, respectively.

We shall verify two key properties: (i) the potential exhibits a low dimensional structure of local minimizers close to natural images, and (ii)
there is a spectral gap similar to the Morse--Bott non-degeneracy condition.
Taken together, these observations suggest that the Morse--Bott framework is a relevant and appropriate surrogate model for the complex,
non-convex geometry of diffusion-based potentials.

\subsubsection{Locating the local minimizers via Gradient Descent}\label{sec:critical_points}
Following the established hypothesis that natural images reside on low-dimensional structures~\cite{popeintrinsic}, we first investigate whether the
diffusion potential \(q\) admits these structures as local minima. We perform gradient descent on the potential \(q\) using the score \(\nabla q\) provided
by the pre-trained diffusion models. The optimization is initialized from three distinct sources: natural images \(x_i\) from the training set, images
from unrelated datasets, and pure white Gaussian noise.

\begin{figure}[htbp]
    \centering
    \subfloat[natural images \(x_i\) from a
    dataset]{\label{fig:crit_point_initial}\includegraphics[width=0.5\linewidth]{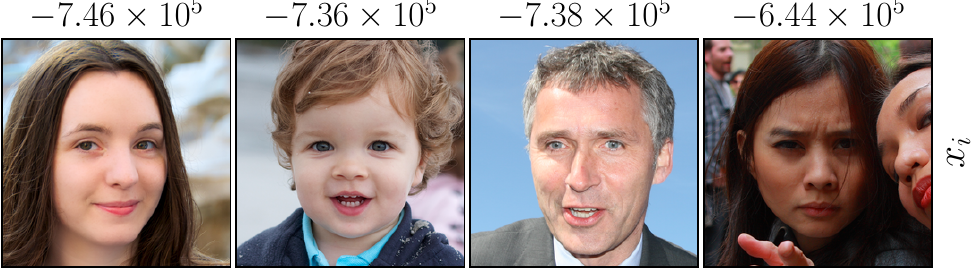}}
    \subfloat[corresponding critical points \(\bar
    x_i\)]{\label{fig:crit_point_end}\includegraphics[width=0.5\linewidth]{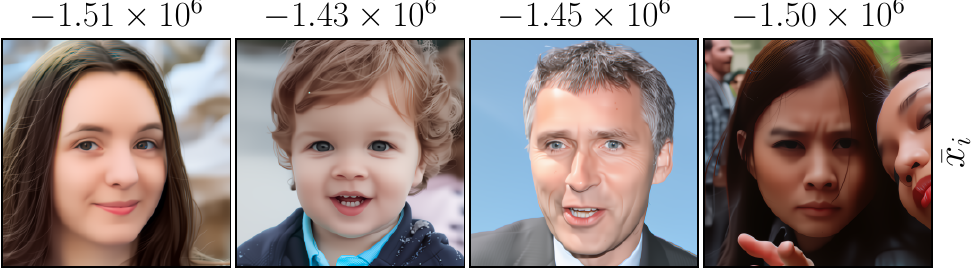}}
    \caption{Critical points obtained by a gradient descent starting from real images in FFHQ-256 using the FFHQ-256 potential. The critical points
        behave as denoised versions of the initial image. The potential \(q=- \log p_{\x}\) displayed on the top of each image is about twice lower for
    critical points.}
    \label{fig:qualitative_crit_points}
\end{figure}

\begin{figure}[htbp]
    \centering
    \includegraphics[width=0.4\linewidth]{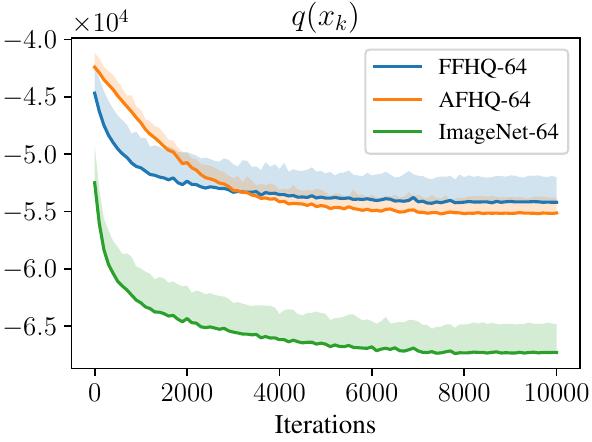}
    \includegraphics[width=0.4\linewidth]{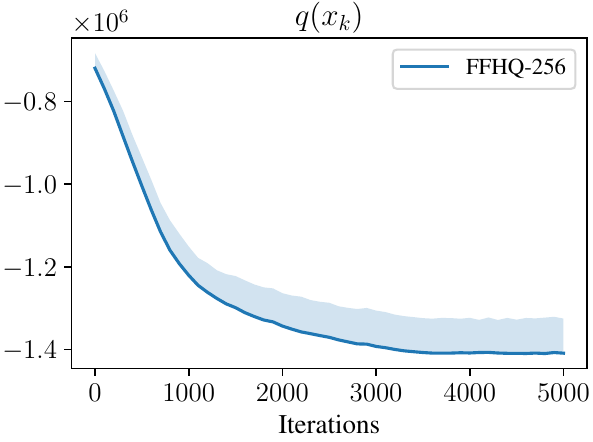}
    \caption{Illustration of the convergence of a gradient descent on the potential \(q\) using different models. The potential stabilizes after a few
        thousands iterations.
        The mean (solid line) and standard deviation (bluish region) across \(100\) different images are displayed.
    }
    \label{fig:cv_gradient_descent_crit}
\end{figure}

Trajectories starting from \(x_i\) reach near-critical points \(\bar{x}_i\) satisfying \(\nabla q(\bar{x}_i) \approx 0\).
Under mild smoothness assumptions, gradient descent avoids strict saddle points almost surely~\cite{lee2016gradient} and converge to second-order
critical points where \(\nabla^2 q(\bar{x}) \succeq 0\).  \Cref{fig:qualitative_crit_points} illustrates that these critical points \(\bar{x}_i\) can be
interpreted as ``denoised'' or ``idealized'' versions of the initial points. The monotonic decrease and stabilization of \(q\)
in~\cref{fig:cv_gradient_descent_crit} suggest that the optimization successfully reaches local minimizers.

\subsubsection{Landscape Curvature}
\begin{figure}[htbp]
    \centering
    \subfloat[FFHQ-256 model]{\label{fig:crit_ffhq256model}\includegraphics[width=0.5\linewidth]{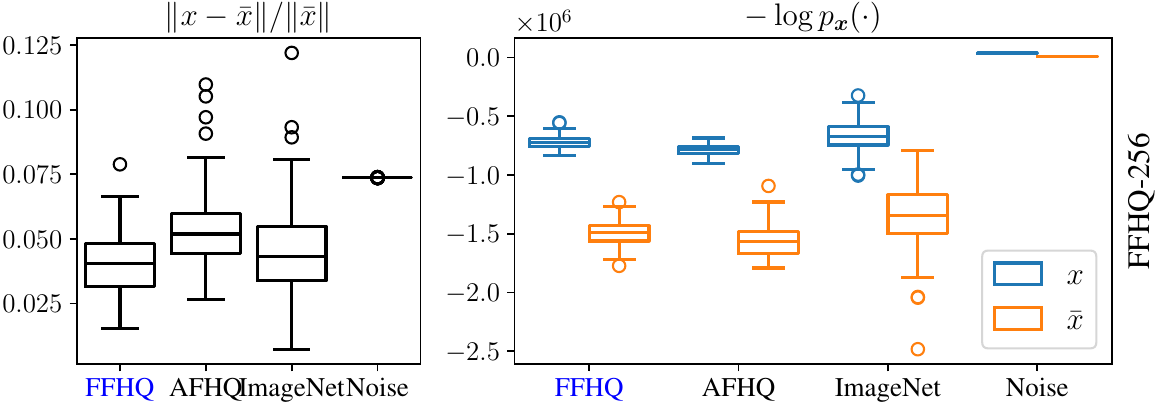}}
    \subfloat[ImageNet-64 model]{\label{fig:crit_imagenet64model}\includegraphics[width=0.5\linewidth]{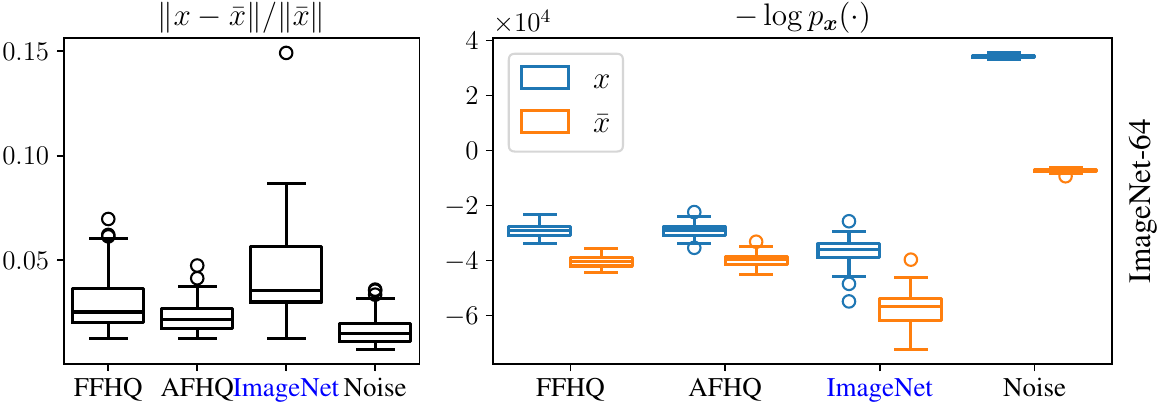}}
    \\
    \subfloat[FFHQ-64 model]{\label{fig:crit_ffhq64model}\includegraphics[width=0.5\linewidth]{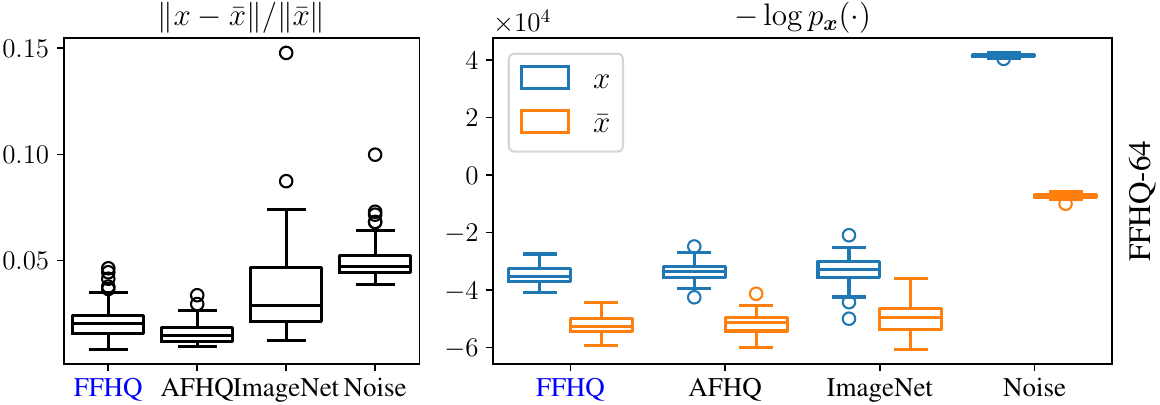}}
    \subfloat[AFHQ-64 model]{\label{fig:crit_afhq64model}\includegraphics[width=0.5\linewidth]{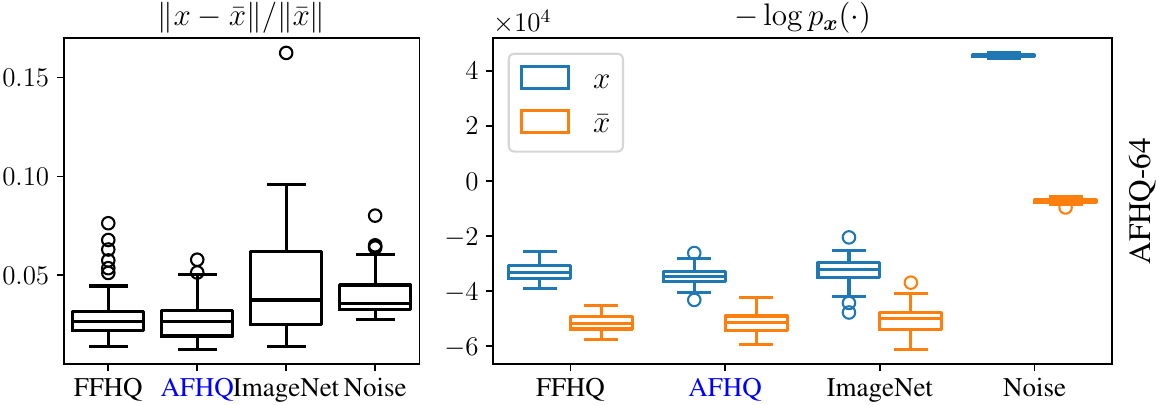}}
    \caption{Distance and difference of potential between natural images and the corresponding critical points.
        The dataset used to train the model is highlighted in blue.
        Observe that the relative distance between the critical points and the initial images is less than 10\% even when starting from pure noise.
        Yet, the potential is much lower.
    }
    \label{fig:crit_boxplot}
\end{figure}

In \cref{fig:crit_boxplot}, we report the relative distance \(\|x_i - \bar{x}_i\| / \|\bar{x}_i\|\) and the corresponding variation in potential. We
observe that a wide variety of initialization, including pure noise, reside in proximity (typically within 10\%) to a critical point. However, the
potential \(q(x_i)\) for noise is orders of magnitude higher than for natural images.

To characterize this behavior locally, let \(d = \frac{x - \bar{x}}{\|x - \bar{x}\|}\) be the unit direction connecting an initialization to its
associated critical point, and define the one-dimensional restriction
\[
    q^{d}(s) = q(\bar{x} + s d).
\]
Since \(\nabla q(\bar{x}) \approx 0\), a second-order Taylor expansion yields
\[
    q(x) - q(\bar{x})
    \;\approx\;
    \frac{1}{2}\|x - \bar{x}\|^{2}
    \langle \nabla^{2} q(\bar{x}) d, d \rangle.
\]
Equivalently, the curvature in direction \(d\) is given by the Rayleigh quotient
\[
    (q^{d})''(0)
    =
    \langle \nabla^{2} q(\bar{x}) d, d \rangle
    \;\approx\;
    \frac{2\big(q(x)-q(\bar{x})\big)}{\|x-\bar{x}\|^{2}}.
\]
Hence, the large increase observed in potential relative to the small distance
\(\|x-\bar{x}\|\) indicates a large curvature in directions transverse to the set of critical points.
This behavior is consistent with a Morse--Bott structure, in which the potential is sharply curved in normal directions while remaining comparatively
flat along the underlying low-dimensional manifold.

% To characterize this locally, let $d = (x - \bar{x})/\|x - \bar{x}\|$ be the direction linking an initialization to its corresponding critical
% point, and let $q^{d}(s) = q(\bar{x} + s \cdot d)$ be the 1D profile of the potential. The curvature in this direction is given by the Rayleigh
% quotient $(q^{d})''(s) = \langle \nabla^2 q(\bar{x}) d, d \rangle$. The observation that $q(x) \gg q(\bar{x})$ for points $x$ near $\bar{x}$
% implies an extremely high curvature in directions leading away from the manifold. This is the first indicator of the Morse--Bott structure: the
% potential blows up in normal directions that do not preserve image-like structures.%, while remaining relatively flat along the manifold itself.

\subsubsection{Hessian Spectral Analysis}
The crucial part of the Morse--Bott property is the existence of a spectral gap in the Hessian \(\nabla^2 q(\bar{x})\) whose kernel identifies the
tangent space \(T_{\bar{x}}\mathcal{M}\).
We compute the full spectrum of the Hessian at several critical points \(\bar{x}_i\). Since \(\nabla q\) is parameterized by a neural network, we
evaluate the Hessian-vector products \(\nabla^2 q(\bar{x}) \cdot e_k\) via automatic differentiation. For a \(3 \times 64 \times 64\) resolution image,
the ambient dimension is \(N = 12288\). We construct the dense \(N \times N\) Hessian matrix as
\begin{equation*}
    \nabla^2 q(\bar{x}) =
    \begin{pmatrix}
        \nabla^2 q(\bar{x}) \cdot e_1 \\
        \vdots \\
        \nabla^2 q(\bar{x}) \cdot e_N
    \end{pmatrix} \in \R^{N \times N},
\end{equation*}
and compute its singular values in double precision.

\begin{figure}[htbp]
    \centering
    \subfloat[FFHQ-64 model]{\label{fig:spectra_ffhq64}\includegraphics[width=0.4\linewidth]{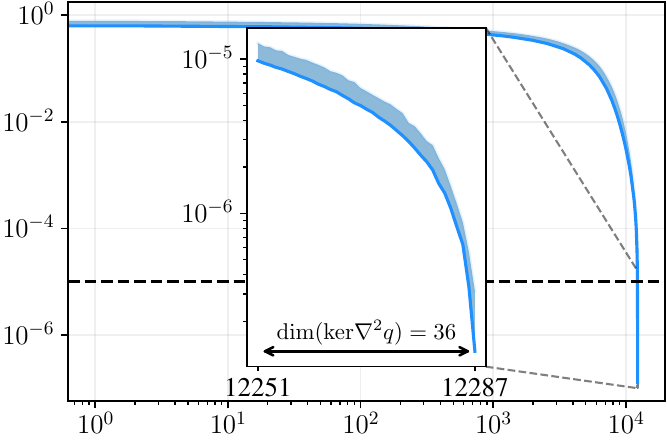}}
    \hspace{1cm}
    \subfloat[AFHQ-64 model]{\label{fig:spectra_afhq64}\includegraphics[width=0.4\linewidth]{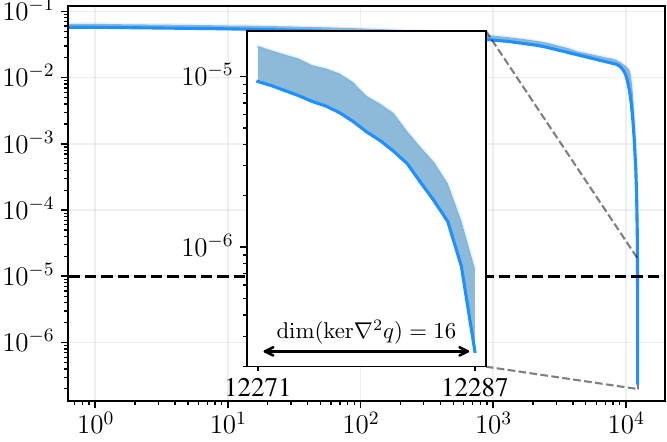}}
    \caption{Spectra of the Hessian of the potential \(\nabla^2 q(\bar x_i)\) for 20 different critical points \(\bar x_i\). By thresholding all values
        below \(10^{-5}\), the local image manifold dimension is estimated at 36 for FFHQ-64 and 16 for AFHQ-64.
        The mean and standard deviation across \(20\) different images are displayed.
        \label{fig:spectra_Hessian}
    }
\end{figure}

The resulting spectra, shown in \cref{fig:spectra_Hessian}, reveal a clear separation of scales. Only a small portion of the eigenvalues are
near-zero, while the vast majority are large and positive. This concentration of small eigenvalues is compatible with the hypothesis of a local image
manifold of small intrinsic dimension \(\mathcal{M}\) \cite{popeintrinsic} and the non-degeneracy condition \(\ker (\nabla^2 q(\bar x)) = T_{\bar x}
\mathcal{M}\). We shall see that this property ensures that the potential acts as a stable anchor for local recovery, despite the global complexities
of the landscape.

\begin{remark}
    Our approach to intrinsic dimensionality via Hessian spectra was carried out independently, and aligns with recent observations by
    Pidstrigach~\cite{pidstrigach2022score} and Stanczuk et al. \cite{stanczuk2024diffusion}, who similarly utilized score-based models to explore
    manifold structures. Consequently, the Morse--Bott property provides a formal theoretical framework which is aligned with these recent empirical
    observations.
\end{remark}

\begin{remark}[On the choice of diffusion potentials]
    While our empirical analysis focuses on diffusion-based potentials, the Morse--Bott framework and the recovery guarantees established in our
    results apply to any \(\mathcal{C}^2\) potential. We choose diffusion models as our primary study case for two reasons:
    (i) they represent the current state-of-the-art in generative modeling of natural images;
    (ii) they provide direct access to the score function \(\nabla q(x)\), enabling a precise numerical mapping of the Hessian spectrum.
    We expect the geometry of the posterior landscape to be a universal feature of high-performance potentials trained on natural image manifolds.
\end{remark}

\begin{remark}[Conservative fields]
    In our experiments, the score \(\nabla q\) is approximated using a pre-trained denoiser via Tweedie's formula~\cite{efron2011tweedie}.
    The \emph{theoretical} score \(\nabla q(x)\) is a conservative vector field (i.e., its Jacobian is symmetric, and it integrates to a scalar potential).
    However, as noted in~\cite{reehorst2018regularization, hurault2022gradient}, a generic neural network is not guaranteed to satisfy this
    integrability condition perfectly, meaning it may not correspond exactly to the gradient of any function.

    While some recent works propose architectures that are conservative by construction~\cite{hurault2022gradient}, we follow the standard approach
    in Score-Based Generative Modeling~\cite{song2021scorebased}: we rely on the fact that the \emph{optimal} MMSE denoiser is inherently
    conservative. Consequently, we treat the non-integrability of the learned network as a negligible approximation error, assuming that the trained
    denoiser is sufficiently close to the true score of the distribution.
    To substantiate this claim, we conducted an experiment over \(20\) critical points \(\bar x\), for which we evaluated the approximated score \(\nabla
    q(\bar x)\) and its Jacobian \(\nabla^2 q(\bar x)\).
    The lack of symmetry of the Jacobian can be measured with \(\frac{\|\nabla^2 q(\bar x)-\nabla^2 q(\bar x)^T\|_F}{\|\nabla^2 q(\bar x)\|_F}\).
    The interquartile range of this quantity across the 20 critical points is \([0.041, 0.047]\), indicating a modest deviation from integrability that
    we treat as a negligible approximation error.
\end{remark}

\section{Recovery Guarantees and Stability for Local Minimizers}
\label{sec:stability}

In this section, we establish the core theoretical result of this paper: the existence and stability of local minimizers in the posterior landscape.
We show that under the Morse--Bott assumption, second order critical points of the potential remain stable anchors for local recovery. We start with
non-blind inverse problems.
\subsection{Non blind inverse problems}

We first recall the notion of second order critical point.
\begin{definition}[Second order critical point]
    A second order critical point of \(q\) is a point \(\bar x\) such that \(\nabla q(\bar x)=0\) and \(\nabla^2 q(\bar x)\succeq 0\).
\end{definition}
These points are the generic limits of gradient descents~\cite{lee2016gradient}.

\begin{theorem}[Recovery guarantees -- non-blind case]
    \label{thm:stability_nonblind}
    Let \(\Lc(x;y) = \frac{1}{2\sigma^2} \|Ax-y\|^2 + q(x)\) and assume that the following conditions are satisfied:
    \begin{enumerate}
        \item \(\bar{x}\) is a ground-truth input with \(A\bar{x} = \bar{y}\),
        \item \(q: \mathbb{R}^N \to \mathbb{R}\) is a \(\mathcal{C}^2\) potential, locally Morse--Bott at \(\bar{x}\) with local critical submanifold
            \(\mathcal{M}\) and \(\bar x\) is a second order critical point of \(q\),
        \item the \emph{identifiability} condition \(T_{\bar{x}}\mathcal{M} \cap \ker A = \{0\}\) holds.
    \end{enumerate}
    Then \(\nabla L(\bar x;\bar y)=0\) and the Hessian \(\bar H \eqdef \frac{1}{\sigma^2}A^T A + \nabla^2 q(\bar x)\) of \(\Lc\) at \(\bar x\) is
    positive definite:
    \[ \bar H \succeq \mu I \qquad \text{for some }\mu>0.\]
    Therefore, there exist neighborhoods \(U\) of \(\bar x\) and \(V\) of \(\bar y\) such that for every \(y\in V\):
    \begin{enumerate}
        \item \(L(\cdot;y)\) admits a unique local minimizer \(x(y)\in U\),
        \item the map \(y\mapsto x(y)\) is \(\mathcal{C}^1\),
        \item the stability estimate holds:
            \[\|x(y)-\bar x\| \le \frac{\|A\|}{\mu \sigma^2}\, \|y-\bar y\| + o(\|y-\bar y\|).\]
    \end{enumerate}
\end{theorem}

The identifiability condition requires the operator \(A\) to be injective when restricted to the tangent space of the image manifold.
It means that two distinct images on the manifold near \(\bar{x}\) cannot produce the same observation.
Because \(T_{\bar{x}}\mathcal{M}\) is possibly low-dimensional, this condition is significantly easier to satisfy than the global injectivity required
in classical inverse problems.
The minimal eigenvalue \(\mu\) in the theorem above is a quantitative measure of this transversality condition.
The larger \(\mu\), the more robust to noise is the local minimizer.
We recall that a \(\mathcal{C}^2\)-function \(f\) is called locally strongly convex at \(\bar x\), if \(H_f(\bar x)\succeq \mu \Id\) with \(\mu > 0\).
Since \(\bar H\) is positive definite, \(L(\cdot;y)\) is locally strongly convex near \(\bar x\). This implies that a gradient descent initialized
sufficiently close to \(\bar x\) will converge to the local minimizer at exponential rate \cite[Thm 1.2.4]{nesterov2013introductory}, implying
stability with respect to initialization.

\begin{remark}[Connections to structured recovery and restricted strong convexity]
    Both \cref{thm:stability_nonblind} and traditional results in compressed sensing~\cite{candes2005decoding, marz2023sampling} provide
        theoretical recovery guarantees for inverse problems under specific geometric conditions. The \emph{identifiability} condition
        \(T_{\bar{x}}\mathcal{M} \cap \ker A = \{0\}\) can be viewed as a manifold instance of \emph{Restricted Strong Convexity}
        (RSC)~\cite{negahban2009unified}, ensuring that the data-fidelity term has sufficient curvature when restricted to the signal's tangent space. 

        The Morse--Bott property complements this by ensuring strict curvature in the directions normal to the manifold.
        Total posterior stability is thus achieved through a geometric duality: the forward operator provides curvature along the \textit{tangent
        bundle}, while the potential provides curvature along the \textit{normal bundle}.

        Unlike polyhedral \(\ell_1\)-theory where the descent cone (and thus the stability constant) depends on discrete active sets and can vary
        abruptly~\cite{marz2023sampling}, our \(\mathcal{C}^2\) framework ensures that the stability constant \(\mu\) (the spectral gap) varies smoothly
	across the image manifold. Actually, the Morse--Bott property is a smooth version of the partial smoothness property \cite{lewis2002active} which
		has been combined with the same injectivity condition to obtain model consistency in sparse modeling \cite{vaiter2017model} as well as local linear
		convergence of nonsmooth optimization methods \cite{liang2014local}.
\end{remark}

% \begin{theorem}[Local Recovery Guarantees for linear inverse problems]
%     \label{th:stabilityLocalMinimaLinear}
%     Let $A \in \mathbb{R}^{M \times N}$ and $q: \mathbb{R}^N \to \mathbb{R}$ be a $\mathcal{C}^2$ prior.
%     Let $\bar{x}$ be a ground-truth input such that $\nabla q(\bar{x}) = 0$, $\nabla^2 q(\bar{x}) \succeq 0$, and $A\bar{x} = \bar{y}$.
%     Assume that $q$ is Morse--Bott at $\bar{x}$ with critical submanifold $\mathcal{M}$.
%     Suppose the following \emph{identifiability} condition holds $T_{\bar{x}}\mathcal{M} \cap \ker A(\bar{\theta}) = \{0\}$.
%     Then there exist $r,\epsilon > 0$ such that for any noise $b$ with $\|b\| \le \epsilon$, there exists a unique strict local minimizer $x^*$ of
% the posterior $\Lc(x)$ in the ball $\mathcal{B}(\bar{x}, r)$. Furthermore, the recovery is stable in the sense that:
% \[ \|x^* - \bar{x}\| = \mathcal{O}(\|b\|). \]
% \end{theorem}

%The identifiability condition requires the operator $A$ to be injective when restricted to the tangent space of the image manifold.

\subsection{Blind inverse problems}

\Cref{thm:stability_nonblind} is a special case of the following result which describes the posterior landscape for blind linear inverse problems.

%         \item \label{cond:identifiabletheta} $\ker J_{\bar{x}}(\bar{\theta}) = \{0\}$.
%         \item \label{cond:identifiablejoint}  $(A(\bar{\theta}) T_{\bar{x}}\mathcal{M}) \cap \mathrm{Im}\ J_{\bar{x}}(\bar{\theta})= \{0\}$.

\begin{theorem}[Stable recovery conditions -- blind case]
    \label{th:stabilityLocalMinima}
    Consider \(\theta \in \R^P \mapsto A(\theta) \in \mathbb{R}^{M \times N}\) a \(\mathcal{C}^2\) parameterization of the linear operator.
    Let \(\Lc(x,\theta;y)\eqdef \frac{1}{2\sigma^2} \|A(\theta)x-y\|^2 + q(x)\) and \(J_x(\theta) = \frac{\partial}{\partial \theta} (A(\theta) x) \in
    \mathbb{R}^{M \times P}\) denote the Jacobian of the measurements w.r.t. \(\theta\). Assume that:
    \begin{itemize}
        \item \((\bar x,\bar\theta)\) is a ground-truth pair with
            \(A(\bar\theta)\bar x=\bar y\),
        \item \(q:\mathbb{R}^N\to\mathbb{R}\) is \(\mathcal{C}^2\) and locally Morse--Bott at \(\bar x\)
            with critical submanifold \(\mathcal M\) and \(\bar x\) is a second order critical point of \(q\).
    \end{itemize}
    Then, \(\nabla_{x,\theta}\Lc(\bar x,\bar\theta;\bar y)=0\) and the Hessian \(\bar H\) of \(L\) at \((\bar x, \bar \theta)\)
    \begin{equation}\label{eq:joint_hessian}
        \bar H \eqdef
        \begin{pmatrix}
            \partial^2_{x, x} \Lc(\bar x,\bar\theta;\bar y) & \partial^2_{x, \theta} \Lc(\bar x,\bar\theta;\bar y) \\ \partial^2_{\theta, x} \Lc(\bar
            x,\bar\theta;\bar y) & \partial^2_{\theta, \theta} \Lc(\bar x,\bar\theta;\bar y)
        \end{pmatrix}
        =
        \begin{pmatrix}
            \bar{H}_{xx} & \bar{H}_{x\theta} \\ \bar{H}_{\theta x} & \bar{H}_{\theta\theta}
        \end{pmatrix}.
    \end{equation}
    is positive definite \emph{if and only if} the following \emph{geometric compatibility} conditions hold:
    \begin{enumerate}
        \item  \label{cond:identifiablex} \emph{Identifiability on the manifold:}
            \(T_{\bar x}\mathcal M \cap \ker A(\bar\theta)=\{0\}\),
        \item \label{cond:identifiabletheta}  \emph{Identifiability of parameters:}
            \(\ker J_{\bar x}(\bar\theta)=\{0\}\),
        \item \label{cond:identifiablejoint} \emph{Manifold-operator decoupling:}
            \((A(\bar\theta)T_{\bar x}\mathcal M)
            \cap \mathrm{Im}\,J_{\bar x}(\bar\theta)=\{0\}\).
    \end{enumerate}
\end{theorem}

\begin{corollary}[Stability bounds]
    \label{th:stability_bounds}
    Under the conditions in~\Cref{th:stabilityLocalMinima}, there exist neighborhoods \(U\) of \((\bar x,\bar\theta)\) and
    \(V\) of \(\bar y\) such that for every \(y\in V\):
    \begin{enumerate}
        \item \(\Lc(\cdot,\cdot;y)\) admits a unique local minimizer \((x(y),\theta(y))\in U\),
        \item the map \(y\mapsto (x(y),\theta(y))\) is \(\mathcal{C}^1\),
        \item the following first-order expansions hold:
            \begin{equation*}
                x(y)-\bar x = \mathrm{Jac}_x(\bar y) (y-\bar y) + o(\|y-\bar y\|) \quad \mbox{and} \quad  \theta(y)-\bar\theta =
                \mathrm{Jac}_\theta(\bar y) (y-\bar y) + o(\|y-\bar y\|).
            \end{equation*}
            The analytical expressions for the Jacobians are given by
            \begin{align}
                \mathrm{Jac}_x(\bar y) &= \frac{1}{\sigma^2}  S_x^{-1} \left(A(\bar\theta)^T-\bar H_{x \theta} \bar H_{\theta \theta}^{-1} J_{\bar
                x}(\bar\theta)^T \right), \label{eq:jac_x} \\
                \mathrm{Jac}_\theta(\bar y) &= - \frac{1}{\sigma^2}  \bar H_{\theta \theta}^{-1} \bar H_{\theta x} S_x^{-1} \left( A(\bar\theta)^T -
                \bar H_{x \theta} \bar H_{\theta \theta}^{-1} J_{\bar x}(\bar\theta)^T  \right) + \frac{1}{\sigma^2} \bar H_{\theta \theta}^{-1}
                J_{\bar x}(\bar\theta)^T, \label{eq:jac_theta}
            \end{align}
            where \(S_x \eqdef \bar H_{xx} - \bar H_{x\theta} \bar H_{\theta\theta}^{-1} \bar H_{\theta x}\) is the Schur complement of \(\bar H\) with
            respect to \(\bar H_{\theta\theta}\).
    \end{enumerate}

\end{corollary}

The proof of \cref{th:stabilityLocalMinima} and \cref{th:stability_bounds}, provided in \cref{sec:proof_stability_local_minima}, leverages the
Morse--Bott non-degeneracy condition, \(\bar x\) being a second-order critical point and the implicit function theorem to show that the joint Hessian
is strictly positive-definite at \((\bar{x}, \bar{\theta})\).
\begin{remark}[Dependence on noise]\label{remark:reconstruction_error}
    The first-order expansions provide a precise characterization of the local stability of the recovery.
    In particular, for bounded noise \(\|y-\bar y\|\le \varepsilon\), we immediately obtain
    \begin{align*}
        \|x(y)-\bar x\| &\le \sigma_{\max}(\mathrm{Jac}_x(\bar y))\, \varepsilon + o(\varepsilon),
        \\
        \|\theta(y)-\bar\theta\| &\le \sigma_{\max}(\mathrm{Jac}_\theta(\bar y))\, \varepsilon + o(\varepsilon)
    \end{align*}
    where \(\sigma_{\max}(\cdot)\) denotes maximal singular value, that coincides with the spectral norm.
    For stochastic noise \(\boldsymbol{b} \sim \mathcal{N}(0,\sigma^2 \Id)\), let \(\y = \bar y + \boldsymbol{b} = A(\bar \theta)\bar x +
    \boldsymbol{b}\) denote the random observation, in which case \(\mathbb{E}\left[ \|\y - \bar y\|^2\right] = M \sigma^2\).
    On the other hand, we can estimate the expected mean squared error (MSE) as:
    \begin{align*}
        \mathbb{E}\left[  \|x(\y)-\bar x\|^2 \right] &\approx  \sigma^2 \mathrm{Tr}\bigl(\mathrm{Jac}_x(\bar y)^T\mathrm{Jac}_x(\bar y)\bigr) =
        \sigma^2 \|\mathrm{Jac}_x(\bar y)\|_F^2, \\
        \mathbb{E}\left[  \|\theta(\y)-\bar\theta\|^2 \right] &\approx  \sigma^2 \mathrm{Tr}\bigl(\mathrm{Jac}_\theta(\bar
        y)^T\mathrm{Jac}_\theta(\bar y)\bigr) =  \sigma^2 \|\mathrm{Jac}_\theta(\bar y)\|_F^2.
    \end{align*}
    Thus, while the spectral norm times the number of measurements gives a worst-case bound, the Frobenius norm provides a more realistic measure of
    the expected recovery error.
    In~\Cref{subsec:geometric_verification}, we will use these analytical expressions to numerically evaluate the stability of various blur families,
    providing a concrete ranking of their identifiability on the image manifold.
\end{remark}

\begin{remark}[On the size of the basin]
    The radius of the neighborhood \(V\) provided by the implicit function theorem
    can in principle be quantified in terms of the local regularity of the
    posterior. More precisely, if the Hessian \(\nabla^2_{x,\theta}\Lc\)
    is locally Lipschitz with constant \(L_H\) near \((\bar x,\bar\theta)\),
    then a valid basin size scales proportionally to \(\mu/L_H\),
    where \(\mu = \lambda_{\min}(\bar H)\).
    Intuitively, this ratio describes how far one can move from the ground truth before the local strong convexity (quantified by \(\mu\)) is destroyed
    by the nonlinearity of the landscape (quantified by \(L_{H}\)).

    We acknowledge that \(L_H\) is computationally inaccessible for high-dimensional neural potentials (third order derivatives).
        However, this formula serves as a qualitative justification for the \textit{numerical mapping} performed later in Section 5.
        Where the Lipschitz constant \(L_H\) cannot be computed, Section 5 provides an empirical measurement of the basin's extent by visualizing the
    posterior's marginalized profile (\Cref{fig:map_noiseless} and \Cref{fig:map_noisy}).
\end{remark}
\paragraph{A Bridge to Non-Convex Recovery}

\Cref{th:stabilityLocalMinima} characterizes a class of images that can be recovered as stable local minima of the posterior. This result extends the
recovery theories developed in compressed sensing \cite{ahmed2013blind, kech2017optimal} to the domain of arbitrary, learned potentials. While
traditional results often require global convexity or range constraints \cite{bora2017compressed}, our Morse--Bott framework accommodates the
complex, non-convex landscapes generated by modern diffusion models.

\subsection{Interpretation of the Geometric Conditions}

The local stability of the inverse problem relies on a geometric compatibility between the image manifold \(\mathcal{M}\) and the forward operator
\(A(\theta)\). While condition \eqref{cond:identifiablex} guarantees that the image \(\bar x\) is uniquely recoverable when the operator is known, the
blind setting in~\cref{th:stabilityLocalMinima} introduces two additional requirements to resolve the joint ambiguity between the signal \(x\) and the
parameters \(\theta\).
We provide below an intuitive interpretation of these conditions, which collectively ensure that the potential \(q\) and the operator \(A\) interact
transversely.

\paragraph{Identifiability of Parameters \eqref{cond:identifiabletheta}}
This is a standard requirement for the identifiability of the forward operator. It ensures that variations in the parameter vector \(\theta\) lead to
measurable changes in the synthesized image.
Without this, there would exist a continuous family of parameters indistinguishable from \(\bar \theta\), even if the ground-truth image \(\bar x\) were
perfectly known.

\paragraph{The Decoupling Condition \eqref{cond:identifiablejoint}}
This condition prevents the ``cross-talk'' between image variations and operator variations. Intuitively, it states that a change in the image along
the manifold cannot be mistaken for a change in the operator's parameters.

To see this, consider a first-order perturbation \((dx, d\theta)\) around \((\bar{x}, \bar{\theta})\). At first order, the synthesized image varies as:
\begin{equation*}
    A(\bar{\theta} + d\theta)(\bar{x} + dx) \approx \bar{y} + \underbrace{A(\bar{\theta}) dx}_{\text{image change}} + \underbrace{J_{\bar
    x}(\bar{\theta}) d\theta}_{\text{operator change}}
\end{equation*}
If the subspaces \(A(\bar{\theta})T_{\bar x}\mathcal M\) and \(\mathrm{Im}\,J_{\bar x}(\bar{\theta})\) had a non-trivial intersection, we could find a
non-zero pair \((dx, d\theta)\) such that these two terms cancel out (\(A dx = - J d\theta\)). In such a scenario, the synthesized image would remain
constant while the potential \(q\) (for \(dx \in T_{\bar{x}}\mathcal{M}\)) remains flat. In such a case, the posterior would contain a ridge of
degenerate solutions, and strict local minimality would be lost. Condition \eqref{cond:identifiablejoint} ensures that the signal subspace and the
operator subspace are \textit{transverse}, guaranteeing that the ground truth is an isolated, stable point in the landscape.
\section{The MAP Paradox: Global Bias and the ``Blurry Trap''}
\label{sec:global_failure}

While \cref{th:stabilityLocalMinima} guarantees that the ground-truth \((\bar{x}, \bar{\theta})\) is a stable local minimizer, it does not ensure that
it is the global one. In fact, a long-standing paradox in blind deconvolution is that global MAP optimization consistently favors a degenerate
``no-blur'' solution, where the operator \(A(\theta)\) collapses to the identity and the image \(x\) remains blurry. This phenomenon, well-known for
handcrafted potentials~\cite{levin2009understanding,benichoux2013fundamental}, is often attributed to the inherent simplicity of sparsity-promoting models.

In this section, we show that this failure persists even with state-of-the-art diffusion potentials. We demonstrate that the ``blurry trap'' is not a
limitation of potential quality, but a fundamental structural property of the learned potential landscape \(q\). Specifically, while the potential
admits local minimizers around natural images, its global landscape is ``slanted'' toward blurry structures, effectively decoupling the global MAP
estimator from the ground truth.

\subsection{Empirical Diagnostic: Blurry Images are More Likely}
\label{sub:blurry_is_more_likely}

\begin{figure}[htbp]
    \centering
    \def \width {0.9\linewidth}
    \includegraphics[width=\width]{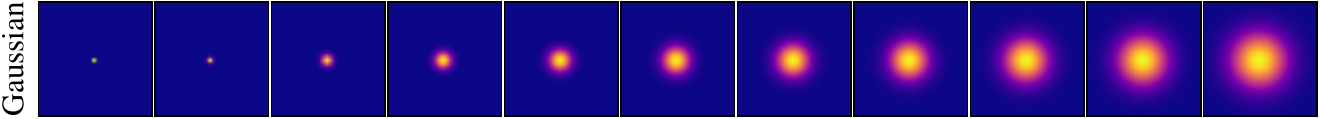}\\
    \includegraphics[width=\width]{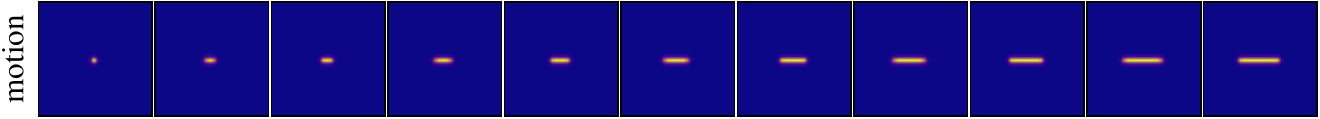}\\
    \includegraphics[width=\width]{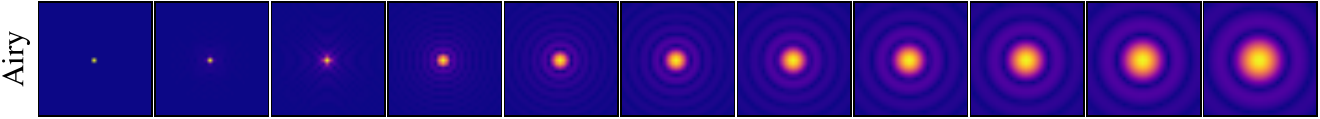}\\
    \includegraphics[width=\width]{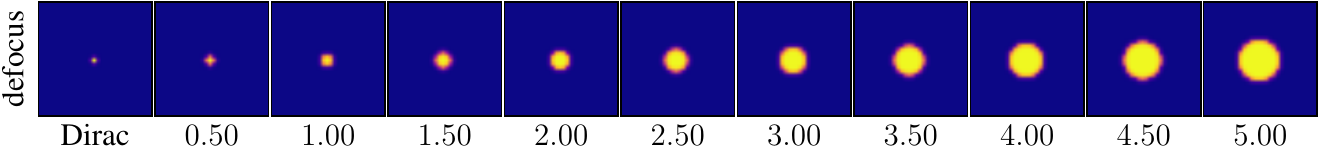}
    \caption{Different 1D blur families used in the forthcoming experiments. The parameter \(\theta\) roughly accounts for the diameter of the PSF with
    \(h_{0} = \delta\) and the higher \(\theta\), the more blur.}
    \label{fig:example_psf_families}
\end{figure}

The failure of the MAP estimate is rooted in a simple but disastrous property: the potential \(q\) tends to decrease as the image becomes blurrier. To
verify this for diffusion-based potentials, we consider a collection of sharp images \(x\) and their blurred counterparts \(h_\theta \star x\), using the
four blur families illustrated in \cref{fig:example_psf_families}.

\begin{figure}[htbp]
    \centering
    \includegraphics[width=0.65\linewidth]{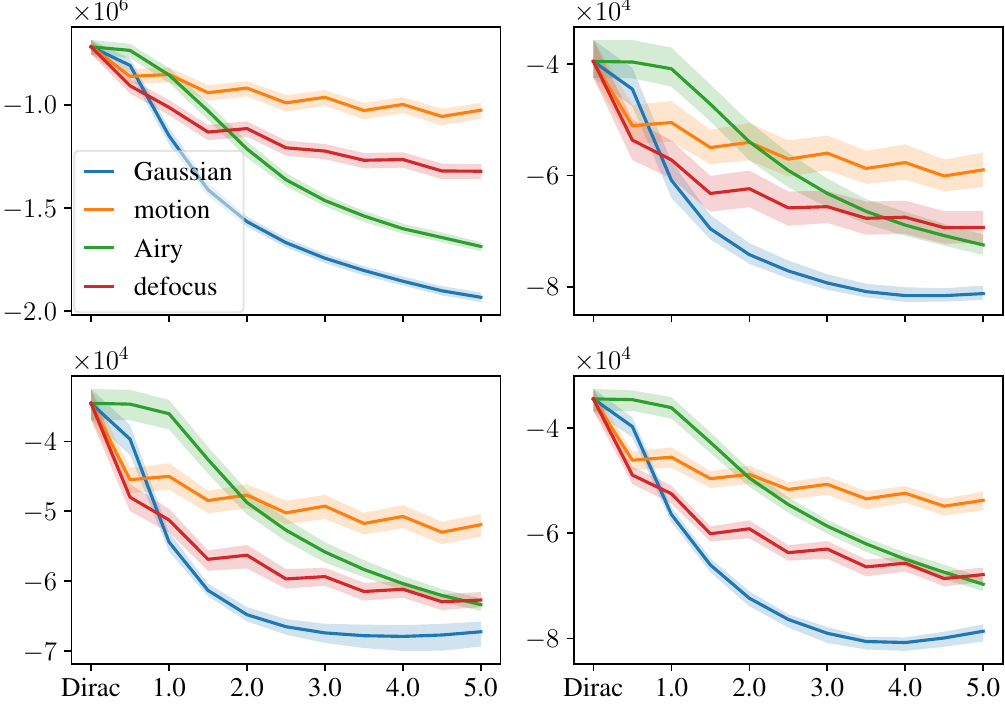}
    \caption{The evolution of the potential \(q(h_{\theta} \conv x)\) on various models and 100 different images, mean and std values are
        shown}. From top-left to bottom-right:
        FFHQ-256, ImageNet-64, FFHQ-64 and AFHQ-64. The initial images are taken from the same training dataset as the pre-trained model. The potential
    is consistently decreasing with the blur level \(\theta\), meaning that blurry images are more likely.
    \label{fig:likelihood_evolution}
\end{figure}

\begin{figure*}[htbp]
    \centering
    \includegraphics[width=\textwidth]{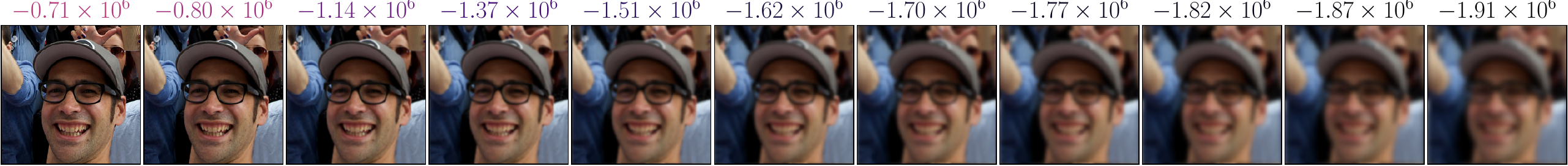}
    \includegraphics[width=\textwidth]{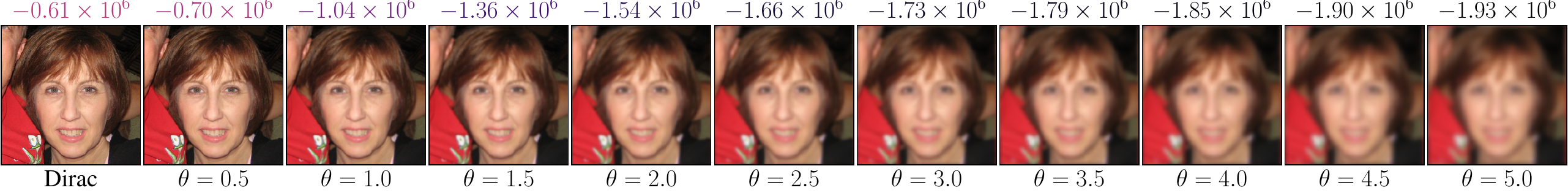}
    \vskip -0.1in
    \caption{Variation of the potential \(q\) with Gaussian blur on the FFHQ-256 model. A smaller potential means a more likely image. Observe that the
    potential is decreasing strictly with the blur level  \(\theta\).}
    %First row: initial images are from the dataset. Second row: initial images are critical points.}
    \label{fig:supp_likelihood_evolution_illustration}
\end{figure*}

\paragraph{Monotonicity of the Potential}
Starting from a sharp image \(x\) from the training set, we evaluate the potential \(q(h_\theta \star x)\) as a function of the blur scale \(\theta\). As
shown in \cref{fig:likelihood_evolution} and qualitatively in~\cref{fig:supp_likelihood_evolution_illustration}, the potential \(q\) decreases strictly
as \(\theta\) increases across all models (FFHQ-256, ImageNet-64, AFHQ-64). This implies that \(p_{\x}(h_\theta \star x) > p_{\x}(x)\), meaning the
potential consistently assigns a higher likelihood to blurred versions of a natural image than to the sharp original.
This behavior aligns with concurrent findings by Karczewski et al.~\cite{karczewski2025diffusion}, who observed that high-density regions of
diffusion models gravitate toward smooth structures. Our results confirm that shifting from handcrafted to learned potentials does not resolve the
preference for blur; it merely encodes this bias into a more complex, non-convex landscape.
\paragraph{A Specificity of Natural Image Statistics}
To determine if this preference for blur is an artifact of neural network architectures (e.g., a bias inherent to the U-Net), we trained an identical
diffusion model on a binarized MNIST dataset. Unlike natural images, binarized digits consist of piecewise constant regions where blur creates
unnatural intermediate intensities.
\begin{figure}[htbp]
    \centering
    \subfloat[True samples]{
        \includegraphics[width=0.25\linewidth]{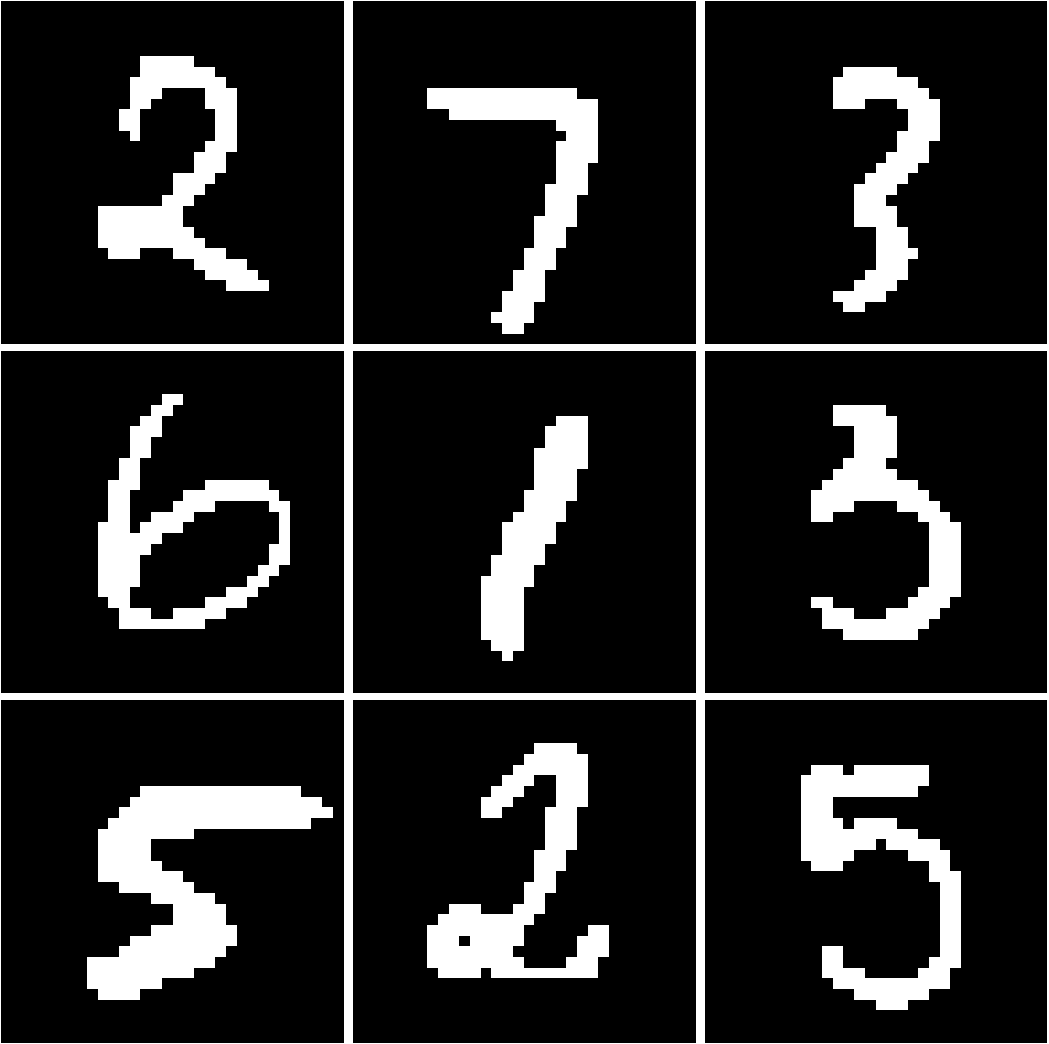}
    }
    \subfloat[Generated samples]{
        \includegraphics[width=0.25\linewidth]{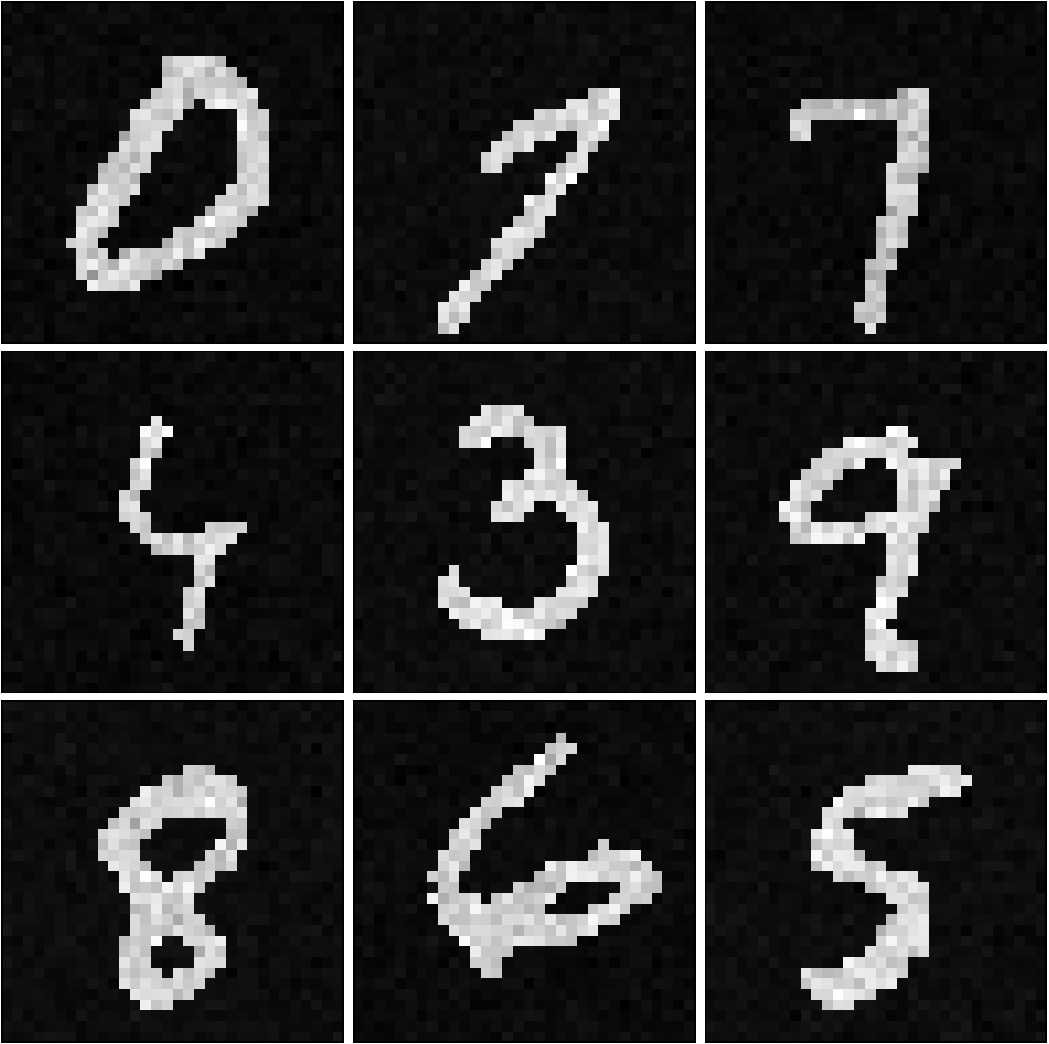}
    }
    \subfloat[Potential evolution]{
        \includegraphics[trim=0cm 0.5cm 0cm 0cm,clip, width=0.4\linewidth]{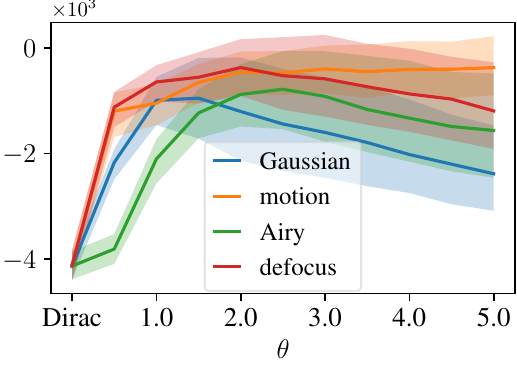}
    }
    \caption{Variation of the potential \(q\) with the blur level, for a model trained on a binarized MNIST dataset. The tendency of the potential to
    decrease with blur is not an architectural artifact, but depends on the statistics of natural images.}
    \label{fig:mnist}
\end{figure}
As shown in \cref{fig:mnist}, the potential for the MNIST model \textit{increases} with the blur level \(\theta\). This experiment suggests that the
``blurry trap'' is not a bias of the neural architecture, but an inherent characteristic of models trained on the statistics of natural images. In
natural photographs, smooth regions and defocus blur are prevalent, leading the learned potential to ``accept'' blurred structures as high-probability states.

\subsection{The Global Failure of MAP}

We now formalize how this global slant in the potential landscape leads to the systematic failure of the MAP estimator. Consider the negative log-posterior:
\begin{equation}\label{eq:def_ly}
    \Lc(x, \theta;y) = \frac{1}{2\sigma^2} \|h_\theta \star x - y \|^2 + q(x).
\end{equation}
In the following theorem, we provide a sufficient condition for the MAP estimator to collapse to the no-blur solution.

\begin{lemma}[Global MAP Failure]
    \label{theorem:no_blur_solution}
    Let \(\mathcal{H} = \{ h_\theta, \theta \in \Theta \}\) be the set of blur kernels, and assume the Dirac delta \(\delta \in \mathcal{H}\). If the
    potential \(q\) satisfies the \emph{blurry preference condition}:
    \begin{equation}\label{eq:keyinequality}
        q(h \star x) \leq q(x) \quad \forall x \in \mathbb{R}^N, \forall h \in \mathcal{H},
    \end{equation}
    then for any observation \(y\), the set of global minimizers in \((x, \theta)\) of \(\Lc(x, \theta;y)\) contains a degenerate pair \((\hat{x}, \delta)\),
    where \(\hat{x}\) is the MAP-denoised version of the blurry observation \(y\).
\end{lemma}

\cref{theorem:no_blur_solution} demonstrates that the ``Blurry Trap'' is not a failure of the potential's generative quality, but a topological
consequence of the potential's global slant.
While \cref{th:stabilityLocalMinima} ensures that the ground truth \((\bar{x}, \bar{\theta})\) is a stable local minimum,
\cref{theorem:no_blur_solution} proves that if the blurry preference~\eqref{eq:keyinequality} holds globally, the ground truth can \textit{never} be
the global maximizer of the posterior.

Our numerical investigations in \cref{sub:blurry_is_more_likely} verify that condition \eqref{eq:keyinequality} is satisfied by state-of-the-art
diffusion potentials. Consequently, blind deconvolution with MAP is characterized by \emph{competing basins of attraction}: a favorable local basin
around the ground truth, and a dominant global basin around the degenerate no-blur solution.

\begin{remark}[Local stability vs. global slant]
    The local stability and the blurry preference (global failure) may coexist in the same landscape.
    Successful recovery thus depends not only on the existence of a solution, but also on an optimization strategy capable of selecting the correct local basin.
    Please refer to~\Cref{sec:numerical_validation} for a numerical experiment that illustrates this co-existence.
\end{remark}

\begin{remark}[Local pits vs. global trend\label{rem:local_pits}]
    Recent work~\cite{karczewski2025diffusion} suggests that diffusion potentials may admit sharp, piece-wise constant (or ``cartoon-like'')
    structures as local potential pits and report evidence showing that  ``Blurring an image increases its likelihood''. This work was conducted
    independently of our own, both came out roughly at the same time, independently observing this behavior. In fact~\cite{karczewski2025diffusion}
    can be seen as an independent confirmation of the preference for blur.
    The preference for blur we report is a \emph{dominant statistical trend}: on average, the potential floor slants toward blurry states. However,
    the potential's landscape is locally non-monotonic and erratic, containing numerous sharp local minimizers with relatively small basins of
    attraction. While these ``pits'' allow the model to generate sharp samples, they do not counteract the global downward pull described
    by~\eqref{eq:keyinequality}, which ultimately drives the global MAP estimator toward the degenerate no-blur solution.
\end{remark}

\subsection{Extension to Other Blind Inverse Problems}
While we focused our experiments on blind deconvolution, the Morse--Bott framework and the ``blurry trap'' are applicable to any blind inverse
problem where the family of operators \(A(\theta)\) exhibits varying spectral decay. For instance, in \emph{blind super-resolution}, the unknown
anti-aliasing kernel plays a role identical to the blur kernel here; a prior preference for smoothness will systematically bias the estimator toward
over-smoothed kernels.
Similarly, in \emph{blind MRI motion correction} or \emph{calibration}, any parameterization that allows the forward mapping to act as a low-pass
filter will likely trigger the same degenerate global MAP behavior.
In all these cases, the identifiability conditions in \cref{th:stabilityLocalMinima} remain the primary tool for guaranteeing that a stable, sharp
solution exists locally, even when it is not the global maximizer. This highlights that the fundamental challenge of blind inverse problems is the
\emph{topological inconsistency} of the MAP estimator.

% ------------------------------------------------------------
% ------------------------------------------------------------
% ------------------------------------------------------------
% ---------                   Practice         ---------------
% ------------------------------------------------------------
% ------------------------------------------------------------
\section{Numerical Experiments}
\label{sec:numerical_validation}
The theoretical duality established in the previous sections suggests that the posterior landscape is characterized by the coexistence of a stable
recovery basin (Section~\ref{sec:stability}) and a dominant, degenerate global attractor (Section~\ref{sec:global_failure}). To visualize this
competition, we explore the joint posterior landscape along the univariate blur scale \(\theta \in [0, \theta_{\max}]\). Specifically, we analyze the
\textit{marginalized} negative log-posterior profile, or the lower envelope of the landscape:
\begin{equation}
    \ell_{y}^{\mathrm{opt}}(\theta) \eqdef \inf_{x \in \mathbb{R}^N} \Lc(x, \theta;y).
\end{equation}
By optimizing over the image \(x\) for a fixed kernel parameter \(\theta\), we reveal the basins of attraction that determine the success or failure of
joint MAP estimation.

\subsection{Noiseless Measurements: The Coexistence of Basins}
We first consider the ideal noiseless scenario, \(\bar{y} = A(\bar{\theta})\bar{x}\), where \(A(\theta) x = h_\theta \star x\).
To facilitate a continuous mapping of the landscape, we ensure the invertibility of the operator for all \(\theta\) by considering slightly perturbed
kernels \(\tilde{h}_\theta = h_\theta + \lambda \delta\), where \(\lambda > 0\) is a small regularization constant. In this regime, the marginalized
profile can be computed explicitly via inverse filtering: \(\hat{x}(\theta) = A(\theta)^{-1}\bar{y}\), which implies
\(\ell_{\bar{y}}^{\mathrm{opt}}(\theta) = q(A(\theta)^{-1}\bar{y})\).
We use the implementation from the DeepInverse library~\cite{tachella2025deepinverse} for efficient inverse filtering.

\begin{figure}[htbp]
    \centering
    \includegraphics[width=0.725\linewidth]{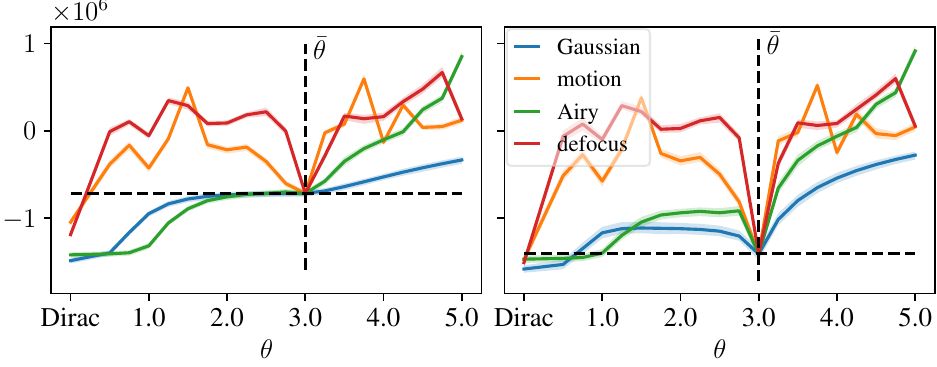}
    \vskip -0.1in
    \caption{Marginalized posterior profiles \(\ell_{\bar y}^{\mathrm{opt}}(\theta)\) for noiseless measurements. Left: \(\bar x\) is a natural image
        from the dataset. Right: \(\bar x\) is a critical point of the potential. The emergence of a local minimum at \(\bar{\theta}\) for critical points
        validates the Morse--Bott stability theory (\cref{th:stabilityLocalMinima}), while the global minimum at \(\theta=0\) validates the global failure
        (\cref{theorem:no_blur_solution}).
    The blur kernel parameterization is the same as in~\cref{fig:example_psf_families}.}
    \label{fig:map_noiseless}
\end{figure}

As shown in \cref{fig:map_noiseless}, the marginalized profiles for the FFHQ-256 model provide a striking visualization of the MAP paradox:
\begin{itemize}
    \item \emph{The Blurry Trap (global failure):} Confirming \cref{theorem:no_blur_solution}, the global minimum of the profile occurs at \(\theta =
        0\) (the Dirac kernel). This demonstrates that the no-blur solution is the global attractor of the landscape, regardless of the prior's
        generative quality.
    \item \emph{The Morse--Bott valley (local stability):} Centering the analysis on a critical point \(\bar{x}\) reveals a distinct local minimum at
        the ground-truth parameter \(\bar{\theta}\). This validates \cref{th:stabilityLocalMinima}, proving that the ``correct'' solution sits within a
        stable, non-degenerate recovery basin.
\end{itemize}

\paragraph{Geometric Identifiability and Fourier Decay}
Interestingly, the profiles for motion and defocus blurs exhibit much sharper local minima than those for Gaussian or Airy blurs. In the context of
\cref{th:stabilityLocalMinima}, this suggests that motion and defocus operators satisfy the \emph{manifold-operator decoupling}
condition~\eqref{cond:identifiablejoint} with higher sensitivity.
It can be seen by smaller values of \(\|\mathrm{Jac}_\theta(\bar y)\|_F\) and \(\sigma_{\mathrm{max}}(\mathrm{Jac}_\theta(\bar y))\)  for the motion
kernel compared to the Gaussian kernel, that we numerically validated in~\Cref{tab:geometric_verification}.
Because the Fourier transforms of these kernels vanish more slowly, the operator \(A(\bar{\theta})\) remains well-conditioned on the tangent space
\(T_{\bar{x}}\mathcal{M}\), leading to a more pronounced basin of attraction and higher stability against perturbations.

\subsection{Numerical Verification of Geometric Identifiability}
\label{subsec:geometric_verification}
The exact recovery guarantees in \cref{th:stabilityLocalMinima} rely on the three geometric transversality
conditions~\ref{cond:identifiablex},~\ref{cond:identifiabletheta} and~\ref{cond:identifiablejoint}, or equivalently, to the invertibility of the
Hessian \(\bar H\)~\eqref{eq:joint_hessian}.
By Schur's formula, the invertibility of \(\bar H\) is equivalent to the invertibility of the Schur complement \(S_x\) and of \(\bar H_{\theta\theta}\).
These conditions are verified in all the experiments below.
However, the mere invertibility of this matrix is qualitative.
We substantiate this invertibility by computing the stability bounds given in~\Cref{remark:reconstruction_error}, which provide more meaningful information.
We evaluate these values using the FFHQ-64 pretrained model. The ground truth images \(\bar{x}\) are taken as critical points of the potential,
obtained as in~\Cref{sec:critical_points}, and we evaluate the stability bounds using different blur kernels displayed in~\cref{tab:geometric_verification}:
\begin{enumerate}
    \item \emph{Isotropic Gaussian blur}: a standard isotropic Gaussian kernel (\(\bar \theta=1.0\), \(P = 1\)), known to be ill-conditioned.
    \item \emph{Motion blur}: a generic motion kernel (length \(\bar\theta = 5, P = 1\)), which preserves more high-frequency information.
    \item \emph{Diffraction blur}: a diffraction blur with \(5\) first aberrations (\(P = 5\), with the same amplitude of \(0.15\) for all aberrations).
    \item \emph{Blur in the simplex}: we also consider a (randomly generated) blur kernel living in the simplex of dimension \(15\times15\) (\(P = 225\)).
\end{enumerate}
We build the Jacobian matrices \(\mathrm{Jac}_x(\bar y)\) and \(\mathrm{Jac}_\theta(\bar y)\) (see~\eqref{eq:jac_x} and~\eqref{eq:jac_theta}), in double
precision floating point, using PyTorch.

Based on~\cref{remark:reconstruction_error}, we then estimate the following quantities:
\begin{align*}
    \varepsilon_{\max, x}(\bar y) &\eqdef \frac{\sigma_{\max}(\mathrm{Jac}_x(\bar y)) \sigma \sqrt{M}}{\|\bar x\|} \qquad \text{and} \qquad
    \varepsilon_{\mathrm{avg}, x}(\bar y) \eqdef \frac{\sigma \|\mathrm{Jac}_x(\bar y)\|_F}{\|\bar x\|} \\
    \varepsilon_{\max, \theta}(\bar y) &\eqdef \frac{\sigma_{\max}(\mathrm{Jac}_\theta(\bar y)) \sigma \sqrt{M}}{\|\bar \theta\|} \qquad  \text{and}
    \qquad \varepsilon_{\mathrm{avg}, \theta}(\bar y) \eqdef \frac{\sigma \|\mathrm{Jac}_\theta(\bar y)\|_F}{\|\bar \theta\|},
\end{align*}
which quantify the worst-case and average relative error in parameter recovery for additive white Gaussian noise \(\boldsymbol{b}\sim
\mathcal{N}(\sigma^2, \Id)\).
The interquartile ranges are displayed in~\cref{tab:geometric_verification} to show the typical range for these values.
\begin{table}[htbp]
    \centering
    \caption{Numerical verification of geometric conditions.
        The interquartile range \([Q_1,Q_3]\) for the maximal and typical relative errors over \(20\) random critical images \(\bar x\) (computed by
        gradient descent, see~\Cref{sec:critical_points}) are presented.
        The parameterization of the PSF has size \(P\) and the noise level is set to \(\sigma = 0.01\). The images are normalized ensuring that \(\bar
        x\in [-1,1]^N\).
        The dimension of the nullspace of \(A(\bar \theta)\) is estimated by thresholding the singular values of \(A(\bar \theta)\) at \(10^{-3}\).
    }
    \label{tab:geometric_verification}
    % \begin{tabularx}{\linewidth}{l *{4}{>{\centering\arraybackslash}X}}
    \resizebox{\linewidth}{!}{
        \begin{tabular}{lcccc}
            \hline
            \multirow{2}{*}{Metric}  & Gaussian & Motion & Diffraction & Blur in the simplex \\
            & (\(P = 1\)) &  (\(P = 1\)) &  (\(P = 5\)) & (\(P = 225\)) \\ \hline
            \raisebox{4\height}{Blur kernel}
            & \includegraphics[width=0.18\linewidth]{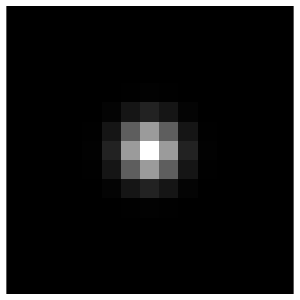}%
            & \includegraphics[width=0.18\linewidth]{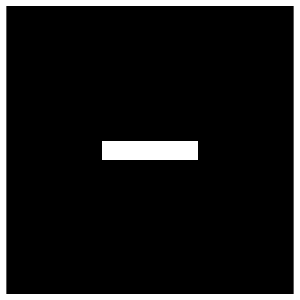}%
            & \includegraphics[width=0.18\linewidth]{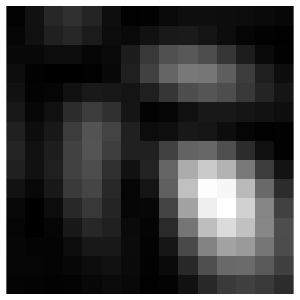}%
            & \includegraphics[width=0.18\linewidth]{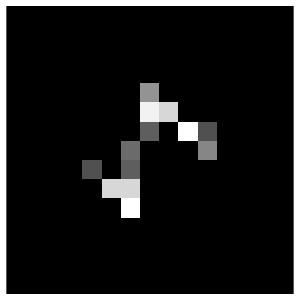}%
            \\ \hline
            \(\mathrm{dim} (\, \mathrm{ker} A(\bar \theta))\) & \(567\) & \(0\) & \(1932\) & \(6\)
            \\\hline
            \(\varepsilon_{\max, x}\)
            & \([0.43, 1.39] \! \times \! 10^{2\phantom{-}}\)
            & \([1.77, 2.93] \! \times \! 10^{0\phantom{-}}\)
            & \([3.07, 9.68] \! \times \! 10^{1\phantom{-}}\)
            & \([1.13, 3.68] \! \times \! 10^{1\phantom{-}}\) \\
            \(\varepsilon_{\max, \theta}\)
            & \([1.27, 2.83] \! \times \! 10^{1\phantom{-}}\)
            & \([3.32, 6.16] \! \times \! 10^{-1}\)
            & \([1.60, 2.53] \! \times \! 10^{1\phantom{-}}\)
            & \([3.52, 11.0] \! \times \! 10^{1\phantom{-}}\) \\
            \hline
            \(\varepsilon_{\mathrm{avg}, x}\)
            & \([2.03, 3.42] \! \times \! 10^{0\phantom{-}}\)
            & \([2.47, 3.37] \! \times \! 10^{-1}\)
            & \([3.96, 6.18] \! \times \! 10^{0\phantom{-}}\)
            & \([2.38, 4.71] \! \times \! 10^{-1}\) \\
            \(\varepsilon_{\mathrm{avg}, \theta}\)
            & \([1.15, 2.55] \! \times \! 10^{-1}\)
            & \([2.99, 5.53] \! \times \! 10^{-3}\)
            & \([2.13, 3.37] \! \times \! 10^{-1}\)
            & \([0.55, 1.45] \! \times \! 10^{0\phantom{-}}\)\\
            \hline
        \end{tabular}
    }
\end{table}
As observed in~\cref{tab:geometric_verification}, the distinction made in~\cref{remark:reconstruction_error} between worst-case bounded noise and
stochastic noise is critical.
For instance, the blur in the simplex presents a high spectral norm bound for the image (\([1.13,3.68] \!\times\! 10^1\)), suggesting severe
vulnerability to adversarial perturbations. However, its expected MSE remains low (\([2.38, 4.71] \! \times \! 10^{-1}\)), indicating that stochastic
noise rarely aligns with the operator's most sensitive singular vectors.

The results also highlight a stark contrast in the conditioning of different blur operators.
The Gaussian kernel and diffraction blur kernels exhibit the most severe ill-posedness, yielding the highest worst-case sensitivity and expected mean
squared error for the image reconstruction.
Conversely, the motion blur preserves more structural high-frequency information, resulting in an expected MSE that is orders of magnitude lower.
Finally, the recovery of the blur parameters \(\theta\) is significantly more stable than that of the image \(x\), except for the motion blur.
This is likely because the parameter dimension \(P\) is significantly smaller than the image dimension \(N\).
For motion blur in the simplex, the kernel lives in a much higher dimension, which may explain a worse identifiability.
In addition, for this case, the fundamental ambiguity in blind deconvolution, where the image and blur are shifted in opposite directions, leaving
the blurry image unchanged, remains.
Yet, the invertibility of the Hessian shows that this ambiguity is mostly solved by the potential.
These experiments overall suggest that the parameter can be recovered rather reliably, while the reconstruction uncertainty on the image \(x\) remains
large: the potential's geometry is not able to compensate for the loss of information.

\subsection{Noisy Measurements and Basin Collapse}
In the presence of noise (\(\sigma > 0\)), the marginalized profile \(\ell_{y}^{\mathrm{opt}}(\theta)\) is no longer explicitly available. We approximate
it by minimizing \(x \mapsto \Lc(x, \theta ; y)\) using a momentum-based gradient descent initialized from multiple points, retaining the endpoint with
the lowest cost value.

\begin{figure}[htbp]
    \centering
    \includegraphics[width=0.75\linewidth]{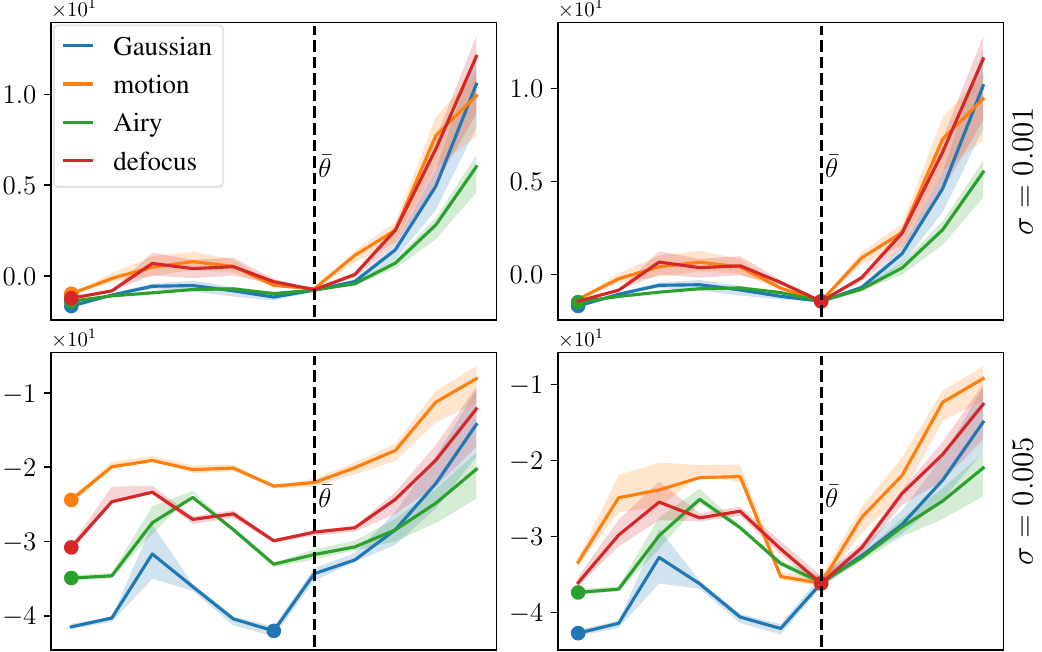}
    \vskip -0.1in
    \caption{Evolution of the marginalized profile \(\ell_{y}^{\mathrm{opt}}(\theta)\) under noisy measurements with natural images (left) and critical
        points (right). As the noise level \(\sigma\) increases, the local recovery basin near \(\bar{\theta}\) vanishes, illustrating the stability limits
        of the Morse--Bott landscape.
    The mean and std values over different images are shown.}
    \label{fig:map_noisy}
\end{figure}

\Cref{fig:map_noisy} illustrates the evolution of the marginalized profile across increasing noise levels. At low noise levels, the posterior
maintains a clear local recovery basin near \(\bar{\theta}\), particularly when the image component is close to a critical point of the potential.
However, as the noise increases, we observe a \emph{basin collapse}: the local recovery minimum is absorbed by the global downward pull of the
``blurry trap.''
This phenomenon suggests the existence of a \emph{critical noise threshold} beyond which the Morse--Bott structure of the recovery basin is
destroyed. For noise levels above this threshold, the ground truth is no longer a local minimizer, and the MAP estimator becomes fundamentally
inconsistent. These results underscore that blind deconvolution is not merely a problem of finding the global minimum, but of ensuring that a
favorable local basin exists and is successfully selected by the optimization algorithm.

% ------------------------------------------------------------
% ------------------------------------------------------------
% ------------------------------------------------------------
% ---------           INITIALIZATION           ---------------
% ------------------------------------------------------------
\section{Navigating the Posterior: Initialization and Optimization}
\label{sec:initialization}

The theoretical and numerical mapping of the posterior in the preceding sections reveals that blind deconvolution is essentially a problem of
\emph{competing attractors}. While the ground-truth pair \((\bar{x}, \bar{\theta})\) resides within a stable Morse--Bott recovery basin, the global
landscape is dominated by a ``blurry trap'' that pulls the optimizer toward a degenerate Dirac solution. Successful recovery thus hinges on a
\emph{basin selection strategy}: the optimization must be initialized within the recovery basin and shielded from the global downward pull.

To navigate this landscape and validate the existence of the recovery basin, we propose a \emph{heuristic} alternating optimization scheme, detailed
in \cref{algo:main_algorithm}.
This solver is not designed to guarantee global convergence, but rather to exploit the topological insights of the Morse--Bott framework using three
guiding principles:

\begin{enumerate}
    \item \emph{Basin-aware initialization:} We initialize \(x_0 = y\) and \(\theta_0\) such that \(h_{\theta_0}\) is the ``maximal'' kernel in the
        parameter space (e.g., maximal spatial support for simplex kernels or high aberration coefficients for Zernike models).
        By starting with a large kernel, we place the initial state in a region where the signal and operator subspaces are transverse, thereby
        satisfying the \emph{manifold-operator decoupling} condition~\eqref{cond:identifiablejoint}.
        When initialized with a small kernel (near the Dirac spike), the data fidelity term \(\|h_\theta\star x -y\|^2\) is minimized by an image \(x\)
        that remains close to the blurry observation \(y\). In this regime, the optimization stays in the ``low-frequency'' part of the image manifold
        \(M\), where the potential \(q\) is most slanted toward the blurry trap. By contrast, a large kernel initialization forces the image component
        \(x\) to contain sharp features and high-frequency content to satisfy the data fidelity.
    \item \emph{Asymmetric alternating optimization:} We perform multiple proximal gradient steps on the image \(x\) for every single gradient step on
        the kernel \(\theta\). This ensures that the image component remains close to the ``manifold floor'' of the Morse--Bott potential, where the
        stability conditions of \cref{th:stabilityLocalMinima} are strictly satisfied.
    \item \emph{Periodic reset (Heuristic escape):} We reset \(x_k = y\) periodically (e.g., every 100 iterations). While this step breaks the strict
        variational structure of the algorithm (disrupting the monotonic decrease of the objective), it effectively ``jumps'' the optimization back
        into the data-consistent space.
        This heuristic serves to reset the data-consistency of the iterate, preventing stagnation in the small, erratic local pits that characterize
        learned potential landscapes (see \cref{rem:local_pits}), allowing it to continue migrating toward the stable basin.
\end{enumerate}

\begin{algorithm}[t]
    \caption{Blind deconvolution with alternating optimization}
    \label{algo:main_algorithm}
    \begin{algorithmic}[1]
        \STATE{\textbf{Input}: blurry image \(y\), number of alternating iterations \(K\), number of iteration for image update \(K_{x}\), step-size
        \(\gamma_x, \gamma_\theta > 0\). }
        \STATE{Initialize \(\theta_0\) and \(x_0 = y\)}
        \FOR{\(k = 0, 1, \dots K - 1\)}
        \STATE{\(x_{k + 1, 0} =
                \begin{cases}
                    y & \textrm{ if } \mathrm{mod}(k, 100) = 0 \\
                    x_{k, K_x} & \textrm{ otherwise}
        \end{cases}\)}
        \\
        \ \textcolor{gray}{\texttt{\textit{Minimization of the posterior with respect to the image}}}
        \FOR{\( i = 0, 1, \dots K_x - 1\)}
        \STATE{\(x_{k + 1, i + 1} = \prox_{\gamma_x f(\cdot, \theta_k)}(x_{k + 1, i} - \gamma_x \nabla q(x_{k + 1, i}))\)}
        \ENDFOR
        \\
        \STATE{\(x_{k + 1} = x_{k + 1, K_x}\)}
        \\
        \ \textcolor{gray}{\texttt{\textit{A gradient descent step for the kernel}}}
        \STATE{\(\theta_{k + 1} = \theta_k - \gamma_\theta \nabla_\theta f(x_{k + 1}, \theta_k)\)}
        \ENDFOR
        \RETURN \(x_K, \theta_K\)
    \end{algorithmic}
\end{algorithm}

In \cref{algo:main_algorithm}, \(f(x, \theta) = \frac{1}{2\sigma^2} \norm{h_\theta \star x - y}^2\) denotes the data fidelity term.  The proximal
operator \(\text{prox}_{\gamma_x f(\cdot, \theta)}\) corresponds to a linear system resolution, which we compute efficiently using the conjugate
gradient method. The gradient \(\nabla q\) is provided by the score of the pre-trained diffusion model, and \(\nabla_{\theta} f\) is computed via
automatic differentiation.

\subsection{Parameterized Blur Kernels: Diffraction-Limited Systems}
We first validate this strategy on diffraction-limited systems, where the PSF \(h_\theta\) is defined by the squared magnitude of the Fourier transform
of the pupil function \(|\mathcal{F}(\exp(-i 2\pi \phi_{\theta}))|^2\). The phase mask \(\phi_\theta\) is parameterized by the first \(P=5\) Zernike
polynomials (defocus, astigmatism, and spherical aberrations) \cite{noll1976zernike}.

\begin{figure}[htbp]
    \centering
    \def\h{0.6\linewidth}
    \begin{minipage}{0.04\linewidth}
        % \raggedright
        \centering
        \rotatebox{90}{\(\bar{x}\)}
    \end{minipage}
    \begin{minipage}{\h}
        \centering
        \includegraphics[width=\linewidth]{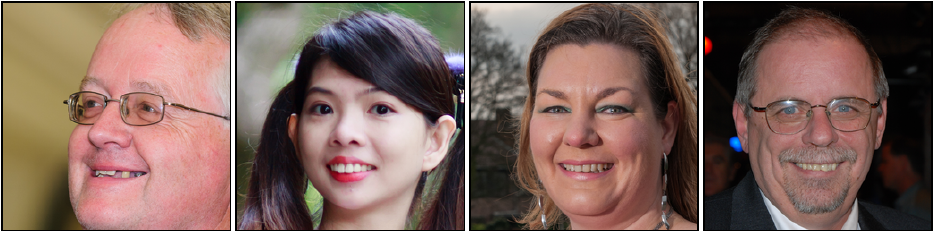}
    \end{minipage}
    \\
    \begin{minipage}{0.04\linewidth}
        % \raggedright
        \centering
        \rotatebox{90}{\(y\)}
    \end{minipage}
    \begin{minipage}{\h}
        \centering
        \includegraphics[width=\linewidth]{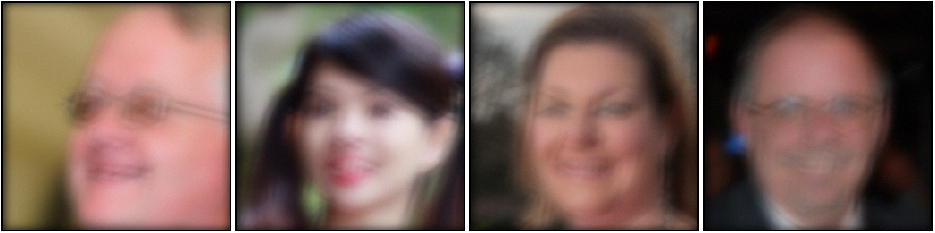}
    \end{minipage}
    \\
    \begin{minipage}{0.04\linewidth}
        % \raggedright
        \centering
        \rotatebox{90}{\(\hat{x}_1\) and \(\hat{h}_1\)}
    \end{minipage}
    \begin{minipage}{\h}
        \centering
        \includegraphics[width=\linewidth]{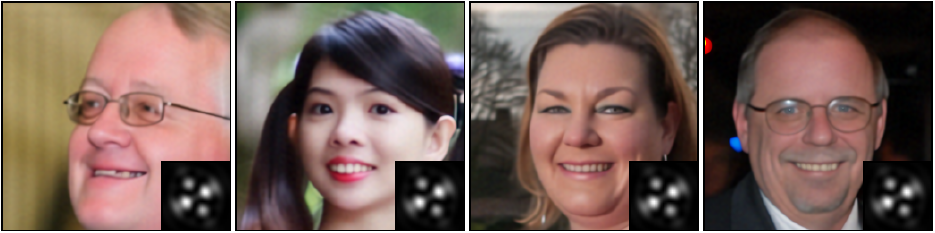}
    \end{minipage}
    \\
    \begin{minipage}{0.04\linewidth}
        % \raggedright
        \centering
        \rotatebox{90}{\(\hat{x}_2\) and \(\hat{h}_2\)}
    \end{minipage}
    \begin{minipage}{\h}
        \centering
        \includegraphics[width=\linewidth]{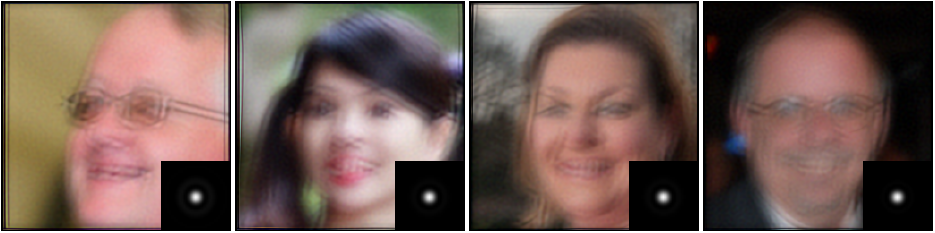}
    \end{minipage}
    \\
    \begin{minipage}{0.04\linewidth}
        % \raggedright
        \centering
        \rotatebox{90}{\(\bar{h}\)}
    \end{minipage}
    \begin{minipage}{0.2\linewidth}
        \centering
        \includegraphics[width=.7\linewidth]{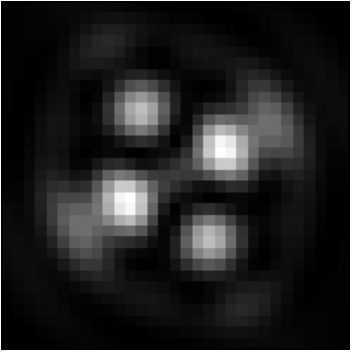}
    \end{minipage}
    \begin{minipage}{0.04\linewidth}
        % \raggedright
        \centering
        \rotatebox{90}{\(h_{\mathrm{init. 1}}\)}
    \end{minipage}
    \begin{minipage}{0.2\linewidth}
        \centering
        \includegraphics[width=.7\linewidth]{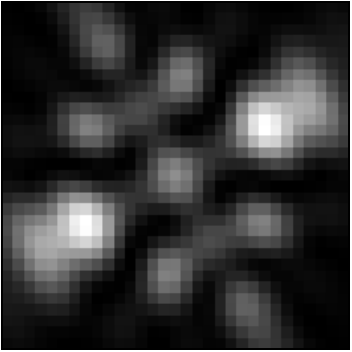}
    \end{minipage}
    \begin{minipage}{0.04\linewidth}
        % \raggedright
        \centering
        \rotatebox{90}{\(h_{\mathrm{init. 2}}\)}
    \end{minipage}
    \begin{minipage}{0.2\linewidth}
        \centering
        \includegraphics[width=.7\linewidth]{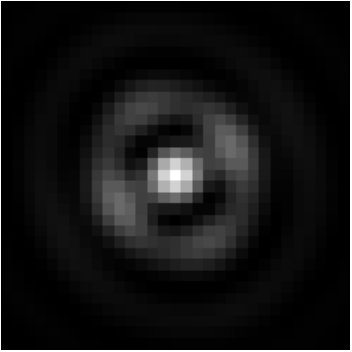}
    \end{minipage}

    \caption{Example of joint posterior minimization for a diffraction-limited blur kernel.
        The \cref{algo:main_algorithm} is initialized at \(y\) (observed with noise level \(\sigma = 0.01\)) and the kernels \(h_{\mathrm{init. 1}}\) or
        \(h_{\mathrm{init. 2}}\).
        The results for \(h_{\mathrm{init. 1}}\) in the 3rd row is satisfactory, while the result for \(h_{\mathrm{init. 2}}\) in the 4-th row is close
        to the no-blur solution.
    This illustrates the importance of initialization when minimizing the posterior jointly. Additional examples are given in \Cref{fig:supp_more_zernike}.}
    \label{fig:zernike_joint_min_larger_kernel}
\end{figure}

In this family, ``larger'' kernels correspond to higher-order aberrations (larger coefficients in \(\theta\)). As shown in
\cref{fig:zernike_joint_min_larger_kernel}, initializing with a complex, large-aberration kernel successfully navigates the landscape to recover the
ground truth. Conversely, starting with a ``simpler'' kernel (closer to the Airy pattern) leads to an immediate collapse into the no-blur solution,
confirming the global pull of the blurry trap.
\subsection{General Blur Kernels in the Simplex}\label{sec:general_kernel}
We further extend the evaluation to general kernels living on the unit simplex, where the mapping \(\theta \mapsto h_\theta\) is the orthogonal
projection onto the simplex. We contrast a ``small'' initialization (a narrow Gaussian) with a ``large'' initialization (a uniform kernel).

\begin{figure}[htbp]
    \centering
    % \rule{\textwidth}{1pt}
    % Grand initialisation
    \def\h{0.0875\linewidth}
    \def\hh{0.7875\linewidth}
    \includegraphics[width=\h]{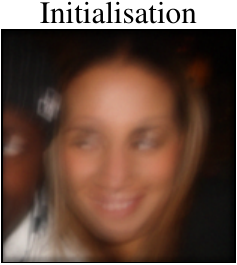}
    \includegraphics[width=\hh, trim={0 0 3.95cm 0}, clip]{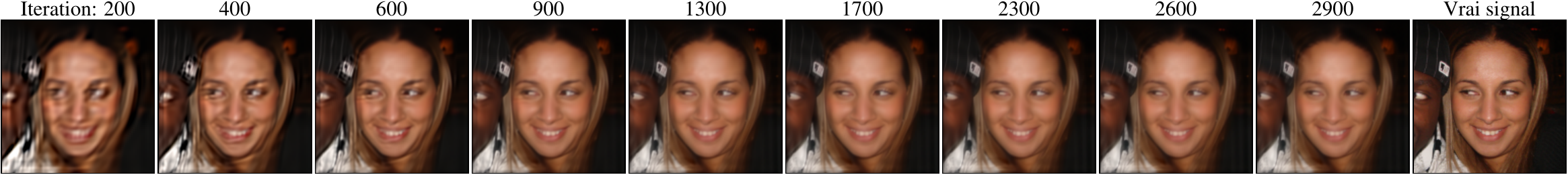}
    \includegraphics[width=\h]{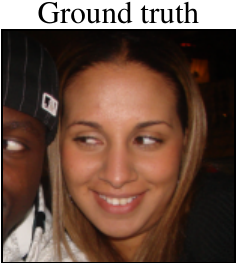}  \\
    \includegraphics[width=\h]{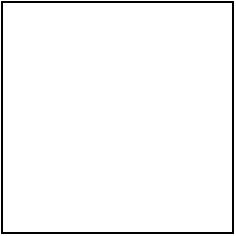}
    \includegraphics[width=\hh, trim={0 0 3.95cm 0}, clip]{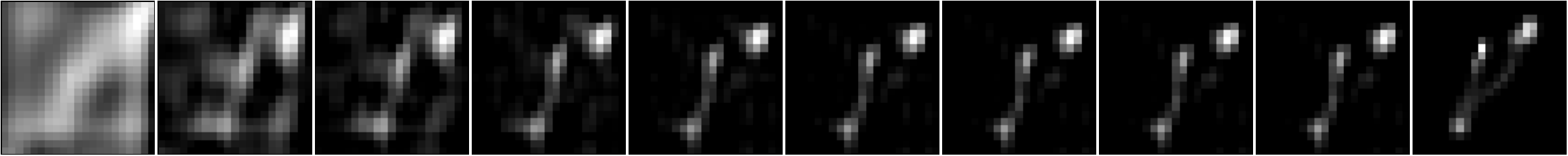}
    \includegraphics[width=\h]{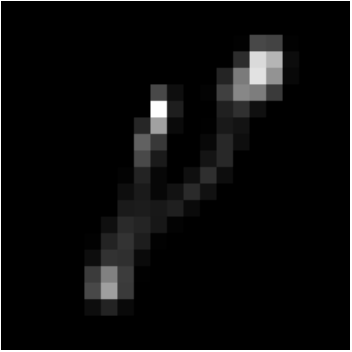} \\

    \includegraphics[width=\h]{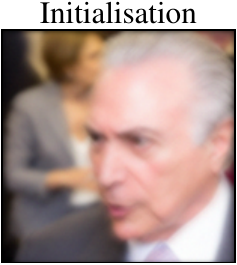}
    \includegraphics[width=\hh, trim={0 0 3.95cm 0}, clip]{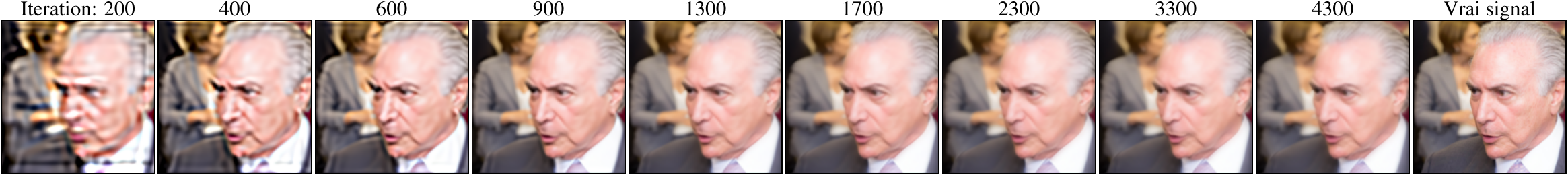}
    \includegraphics[width=\h]{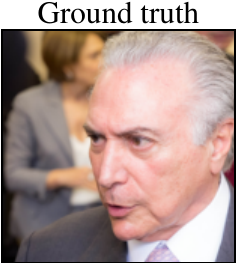}\\
    \includegraphics[width=\h]{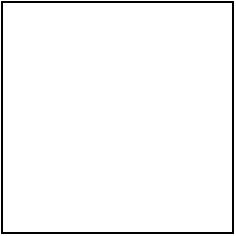}
    \includegraphics[width=\hh, trim={0 0 3.95cm 0}, clip]{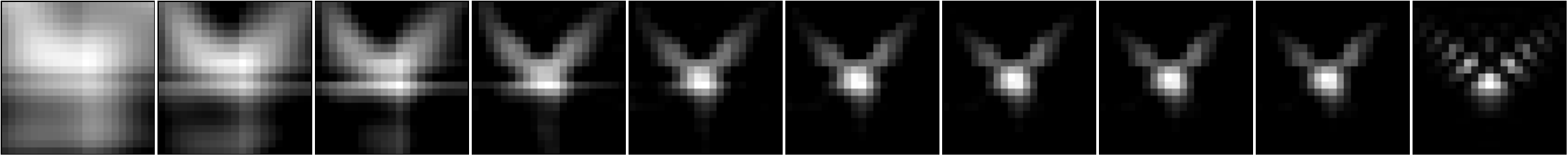}
    \includegraphics[width=\h]{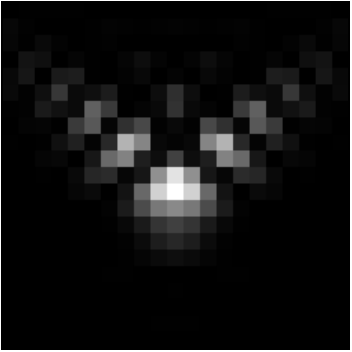} \\
    % Petit initialisation
    \includegraphics[width=\h]{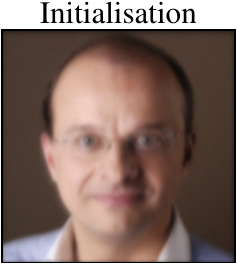}
    \includegraphics[width=\hh, trim={0 0 3.95cm 0}, clip]{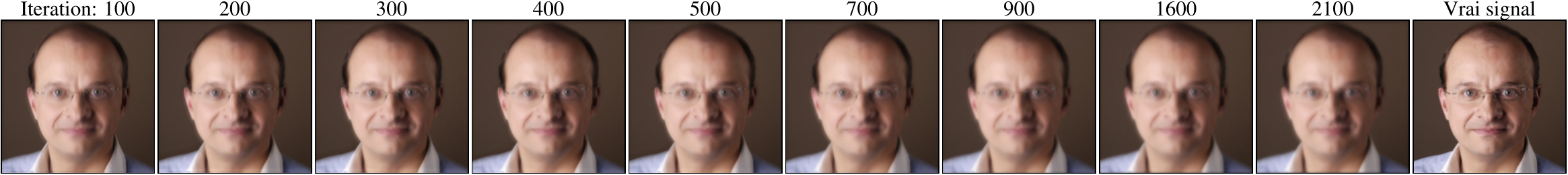}
    \includegraphics[width=\h]{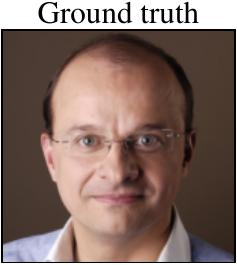}\\
    \includegraphics[width=\h, ]{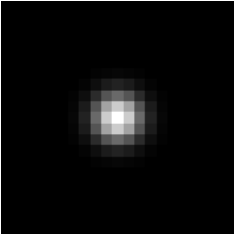}
    \includegraphics[width=\hh, trim={0 0 3.95cm 0}, clip]{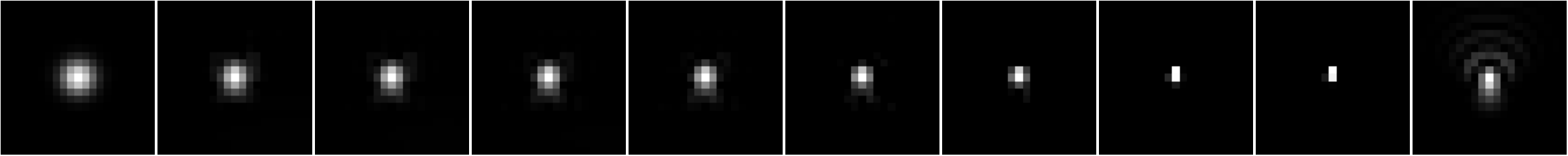}
    \includegraphics[width=\h]{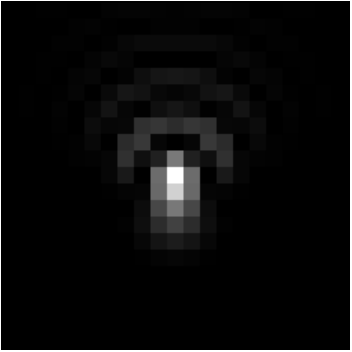} \\

    \includegraphics[width=\h]{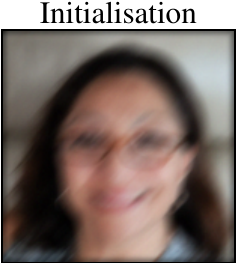}
    \includegraphics[width=\hh, trim={0 0 3.95cm 0}, clip]{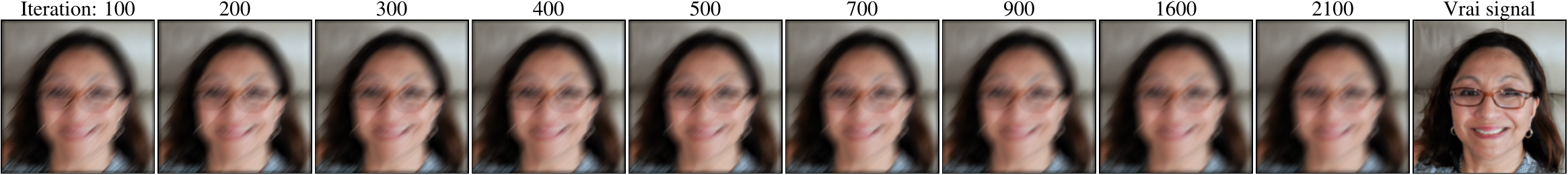}
    \includegraphics[width=\h]{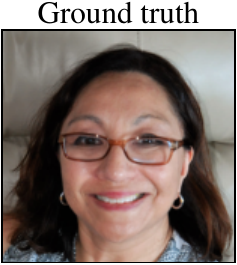}\\
    \includegraphics[width=\h]{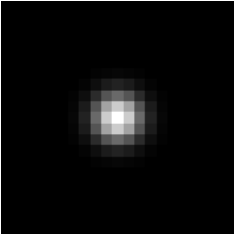}
    \includegraphics[width=\hh, trim={0 0 3.95cm 0}, clip]{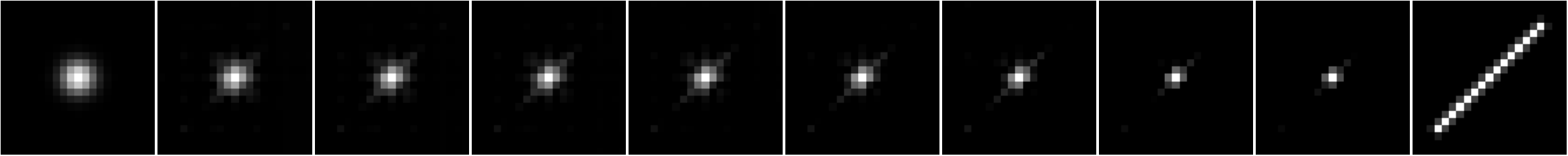}
    \includegraphics[width=\h]{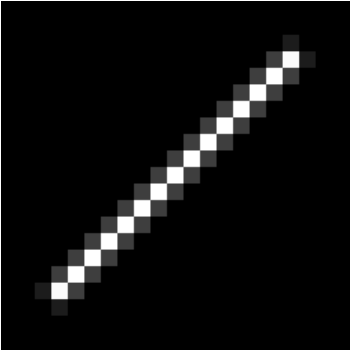} \\

    \caption{Illustration of the \cref{algo:main_algorithm} and the influence of the kernel initilization \(h(\hat{\theta}_0)\).
        A ``large'' initialization (uniform kernel, top two rows) of the kernel results in a satisfactory recovery of the image and the kernel.
        A ``small'' initialization (a Gaussian kernel with standard-deviation of \(1.5\), bottom two rows) of the kernel results in a convergence to
        the no-blur solution.
    We use the FFHQ-256 model from~\cite{song2021scorebased}. }
    \label{fig:numerical_result_simulated}
\end{figure}

\paragraph{Synthetic validation}
Results on four distinct kernels in~\cref{fig:numerical_result_simulated} confirm our theoretical predictions: the small initialization is captured
by the Dirac spike, while the uniform initialization, representing the maximal-entropy state furthest from \(\theta=0\), successfully enters the
recovery basin.
Further results are provided in \cref{fig:supp_more_zernike} for diffraction-limited kernels and in \cref{fig:numerical_result_gretsi_endpoints} for
simplex kernels, demonstrating the consistent efficacy of the proposed strategy across various blur types and image models.

\paragraph{Real-world data: the K\"ohler dataset}
To demonstrate robustness beyond exact training distributions, we evaluate the strategy on the K\"ohler dataset \cite{kohler2012recording} using a
pretrained DRUNet denoiser \cite{zhang2021plug} as a potential. Following the Plug-and-Play (PnP) framework \cite{nguyen:pnp_unrolled}, the denoiser
can be related to the gradient of an implicit potential. This is justified by the score-matching perspective and Tweedie's formula, where the
denoiser output approximates a step along the manifold's gradient \(\nabla q\).

\begin{figure}[htbp]
    \def\size{0.325\linewidth}
    \def\ksize{0.1\linewidth}
    \def\evolsize{0.0875\linewidth}
    \centering
    % -----------------------------TRUE-----------------------------

    \begin{overpic}[width=\size]{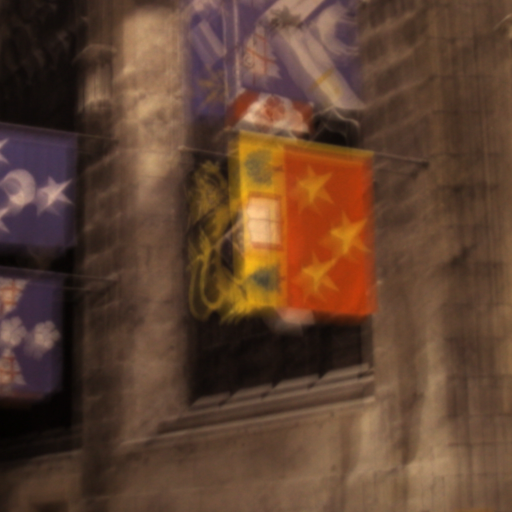}
        \put(0,\size - \ksize){\includegraphics[width=\ksize]{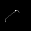}}
    \end{overpic}
    \begin{overpic}[width=\size]{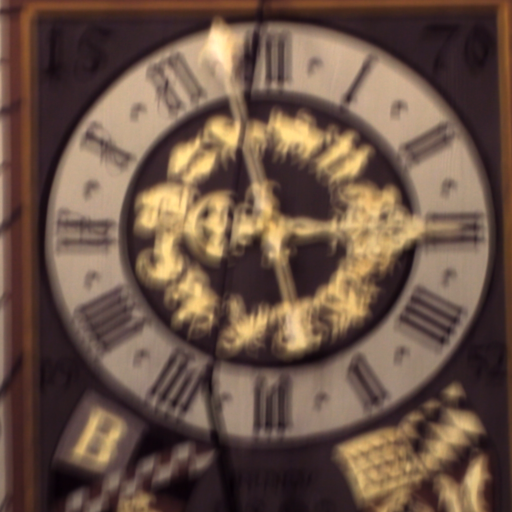}
        \put(0,\size - \ksize){\includegraphics[width=\ksize]{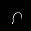}}
    \end{overpic}
    \begin{overpic}[width=\size]{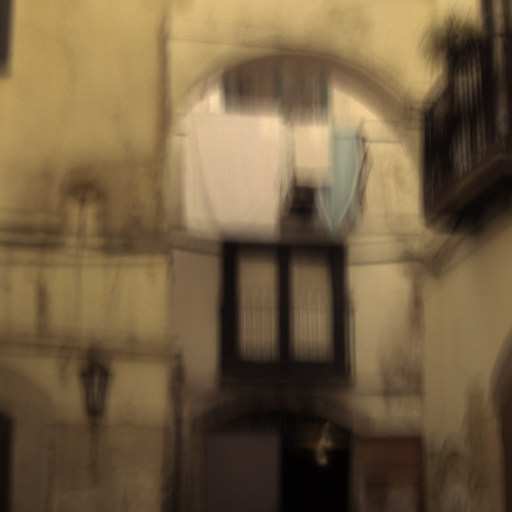}
        \put(0,\size - \ksize){\includegraphics[width=\ksize]{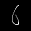}}
    \end{overpic}
    \\
    \vspace{0.15cm}
    % -----------------------------ESTIMATED-----------------------------
    \begin{overpic}[width=\size]{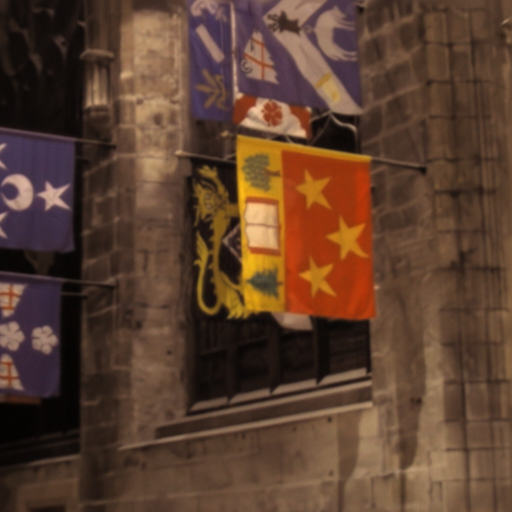}
        \put(0,\size - \ksize){\includegraphics[width=\ksize]{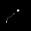}}
    \end{overpic}
    \begin{overpic}[width=\size]{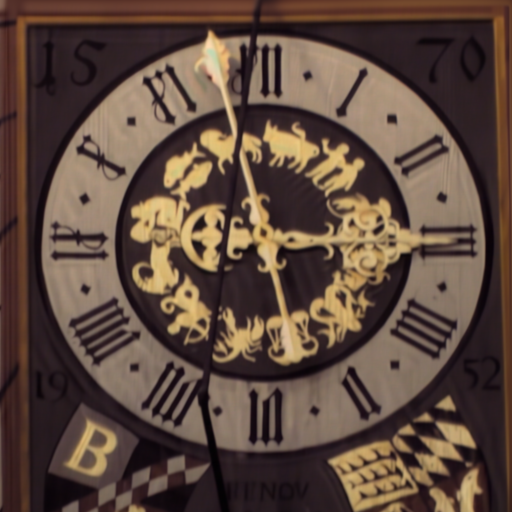}
        \put(0,\size - \ksize){\includegraphics[width=\ksize]{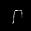}}
    \end{overpic}
    \begin{overpic}[width=\size]{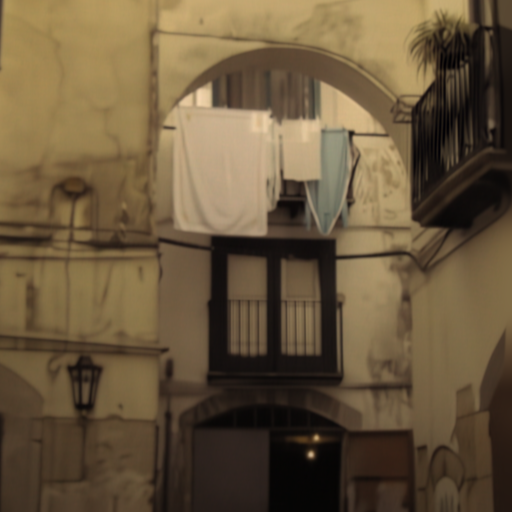}
        \put(0,\size - \ksize){\includegraphics[width=\ksize]{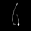}}
    \end{overpic}
    % -------------------------- KERNEL 1 ---------------------------------------------
    \includegraphics[width=\evolsize]{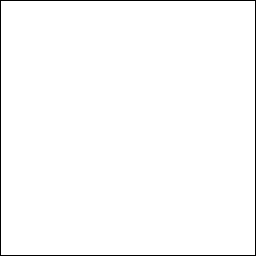}\hspace{-0.9mm}
    \includegraphics[width=\evolsize]{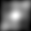}\hspace{-0.9mm}
    \includegraphics[width=\evolsize]{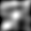}\hspace{-0.9mm}
    \includegraphics[width=\evolsize]{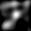}\hspace{-0.9mm}
    \includegraphics[width=\evolsize]{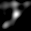}\hspace{-0.9mm}
    \includegraphics[width=\evolsize]{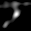}\hspace{-0.9mm}
    \includegraphics[width=\evolsize]{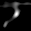}\hspace{-0.9mm}
    \includegraphics[width=\evolsize]{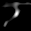}\hspace{-0.9mm}
    \includegraphics[width=\evolsize]{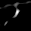}\hspace{-0.9mm}
    \includegraphics[width=\evolsize]{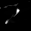}\hspace{-0.9mm}
    \includegraphics[width=\evolsize]{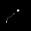}
    % -------------------------- KERNEL 2 ---------------------------------------------
    \includegraphics[width=\evolsize]{./Images/final_result_kohler/kernel_evolution_3/kernel_5_iter_0.png}\hspace{-0.9mm}
    \includegraphics[width=\evolsize]{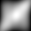}\hspace{-0.9mm}
    \includegraphics[width=\evolsize]{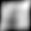}\hspace{-0.9mm}
    \includegraphics[width=\evolsize]{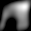}\hspace{-0.9mm}
    \includegraphics[width=\evolsize]{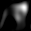}\hspace{-0.9mm}
    \includegraphics[width=\evolsize]{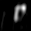}\hspace{-0.9mm}
    \includegraphics[width=\evolsize]{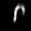}\hspace{-0.9mm}
    \includegraphics[width=\evolsize]{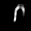}\hspace{-0.9mm}
    \includegraphics[width=\evolsize]{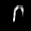}\hspace{-0.9mm}
    \includegraphics[width=\evolsize]{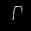}\hspace{-0.9mm}
    \includegraphics[width=\evolsize]{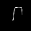}
    % -------------------------- KERNEL 3 ---------------------------------------------
    \includegraphics[width=\evolsize]{./Images/final_result_kohler/kernel_evolution_3/kernel_5_iter_0.png}\hspace{-0.9mm}
    \includegraphics[width=\evolsize]{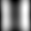}\hspace{-0.9mm}
    \includegraphics[width=\evolsize]{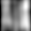}\hspace{-0.9mm}
    \includegraphics[width=\evolsize]{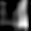}\hspace{-0.9mm}
    \includegraphics[width=\evolsize]{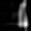}\hspace{-0.9mm}
    \includegraphics[width=\evolsize]{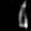}\hspace{-0.9mm}
    \includegraphics[width=\evolsize]{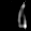}\hspace{-0.9mm}
    \includegraphics[width=\evolsize]{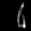}\hspace{-0.9mm}
    \includegraphics[width=\evolsize]{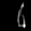}\hspace{-0.9mm}
    \includegraphics[width=\evolsize]{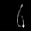}\hspace{-0.9mm}
    \includegraphics[width=\evolsize]{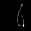}

    \caption{Results on the K\"ohler dataset~\cite{kohler2012recording}. Top: real blurry images and true kernels. Middle: deblurred images and
    estimated kernels. Bottom: estimated kernel along iterations.}
    \label{fig:kohler}
\end{figure}

By initializing with a \(51 \times 51\) uniform kernel and applying \cref{algo:main_algorithm}, we recover high-fidelity images and accurate motion
trajectories (\cref{fig:kohler}). This success on real data, achieved with a generic denoiser, highlights the universal nature of the Morse--Bott
recovery basin and the practical efficacy of topologically informed initialization.

% ------------------------------------------------------------
% ------------------------------------------------------------
% ------------------------------------------------------------
% ---------               CONCLUSION           ---------------
% ------------------------------------------------------------
\section{Conclusion}\label{sec:conclusion}

Our most significant positive result is the general recovery guarantee provided in \cref{th:stabilityLocalMinima}. We establish that for a broad
class of blind inverse problems \(y = A(\theta)x + b\), local recovery is not only possible but stable, provided the operator and the manifold satisfy
three geometric conditions: identifiability on the manifold, parameter identifiability, and manifold-operator decoupling. This framework offers a
rigorous template for analyzing a wide variety of blind problems beyond deconvolution, such as blind super-resolution or blind calibration in
medical imaging.

However, we also demonstrate that this local stability is often overshadowed by a global spectral bias.
% In the case of blind deconvolution, we uncover a ``blurry trap'' where the prior's inherent preference for low-frequency content—a property shared
% by state-of-the-art diffusion models—drives the global MAP estimator toward degenerate solutions. This suggests that for any family of operators
% characterized by varying Fourier decay, the global MAP estimator is fundamentally unreliable.
In the case of blind deconvolution, we demonstrate that the ``blurry trap''--a phenomenon well-documented for sparse
priors~\cite{levin2009understanding}--persists even with state-of-the-art diffusion models. We show that the learned potential's inherent preference
for low-frequency content drives the global MAP estimator toward degenerate solutions, proving that higher generative quality does not resolve the
geometric degeneracy rooted in the forward operator. This suggests that for any family of operators characterized by varying Fourier decay, the
global MAP estimator remains fundamentally unreliable.

Ultimately, our findings suggest a shift in the focus of blind inverse problem research: from the search for global maximizers to the
    characterization and targeting of local recovery basins.
    We conclude that successful recovery in high-dimensional non-convex landscapes is largely a matter of \emph{topologically informed optimization}.
    While we empirically demonstrate that strategic initialization can effectively guide the system into the Morse--Bott recovery basin and bypass the global traps
    of natural image statistics, our work represents only a first step toward a complete theory of navigation in these landscapes. Designing more robust
path-finding heuristics and providing formal certificates for such initialization strategies remain crucial open avenues for future research.

\section*{Acknowledgments}
The authors acknowledge a support from the ANR Micro-Blind (ANR-21-CE48-0008) and from the ANR CLEAR-Microscopy (ANR-25-CE45-3780).
This work was performed using HPC resources from GENCI-IDRIS (Grant AD011012210).

% \null
% \pagebreak
% \null
% \pagebreak
% \null
% \pagebreak
% \null
% \pagebreak
% \null
% \pagebreak
\null
\newpage
\appendix
% ------------------------------------------------------------
% ------------------------------------------------------------
% ------------------------------------------------------------
% ---------               APPENDIX             ---------------
% ------------------------------------------------------------
\section{Details on the experimental setup}\label{sec:experimental_setup}

\paragraph{Diffusion models}\label{sec:diffusion_models}
In our experiments, we focus on state-of-the-art potentials based on diffusion models.
In what follows, bold fonts are used to denote random vectors.
Diffusion models are powerful generative models and based on the following stochastic differential equation (SDE) on \(\R^d\)~\cite{song2021scorebased}:
\begin{equation}
    \label{eq:forward_sde}
    d\x_t = f(\x_t, t) \,dt + g(t)\,d\w_t,
\end{equation}
where \(\w_t\) is the standard Brownian process, \(f\colon \R^d \times \R \to \R^d\) denotes a drift term and \(g\colon \R \to \R\) denotes a
diffusion coefficient.
The choice of \(f\) and \(g\) define the diffusion process, with various qualitative properties~\cite{song2021scorebased,Karras2022edm}.
In this model, \(\x_0\) follows the clean image distribution \(p_{\x}\). For long time horizon \(T\) and under technical conditions, the solution of the
SDE \(\x_T\) approximately follows a normal distribution with no information left on \(p_{\x}\), whose parameters depend on \(f\) and \(g\).
This process can be reversed~\cite{ANDERSON1982313} using the following SDE running backward in time, which serves as a generative process:
\begin{equation}
    \label{eq:backward_sde}
    d\x_t = \left[ f(\x_t, t) - g(t)^2 \nabla \log p_t(\x_t) \right] \,dt + g(t) \,d \w_t.
\end{equation}
Here \(p_t\) denotes the density at time \(t\) and the solutions to this reversed-time SDE are distributed according to \(p_{\x}\) at time \(t=0\).
To generate new samples, we start from \(\x_T \sim \Normal{0, \Id}\) and solve the backward SDE \cref{eq:backward_sde} for \(t \in [T, 0]\) by a
discretization scheme, \eg Euler-Maruyama.
Training a diffusion model amounts to estimating the Stein's score function of each marginal distribution, \(\nabla \log p_t\), by score
matching~\cite{vincent2011connection}.
The score \(\nabla \log p_t\) is typically parameterized by a time-dependent neural network \(s_{\theta}(\cdot, t)\).
For standard diffusion processes and any \(t>0\), \(p_t\) corresponds to a Gaussian mixture, whose variance increases with \(t\). In particular, for any
\(t>0\), \(p_t\) is smooth and strictly positive. In contrast, the data distribution \(p_0\) maybe singular (e.g. supported on a low-dimensional manifold),
making \(\log p_0\) and its gradient ill-defined.

In what follows, we let \(q_t \eqdef -\log p_{t}\) denote the potential (negative-log-prior) term.
We evaluate potentials with a small value of time \(t_\epsilon=10^{-3}\). This value was chosen to obtain a sufficiently smooth approximation of \(p_0\)
compatible with gradient and Hessian computation, while still providing a good approximation of the prior \(p_0\).
In what follows, we simply write \(q\) to denote \(-\log p_{t_\epsilon}\), and we subsequently use \(\nabla q(\cdot)\) to describe the action of the neural
network \(-s_\theta(\cdot, t_\epsilon)\).
This quantity is proportional to a difference between a noisy image and its denoised version via Tweedie's formula~\cite{vincent2011connection}.
\paragraph{Likelihood computation}
There exists a \emph{deterministic process} with trajectories sharing the same marginal distribution as the SDE in \eqref{eq:forward_sde}.
It can be obtained by the Fokker-Plank equation and it is defined by the following ordinary differential equation
(ODE)~\cite{song2021scorebased,Karras2022edm}:
\begin{equation}\label{eq:ode_flow}
    d\x_t =  v(\x_t, t) \,dt, \quad \mbox{ where } v(\x_t, t) \eqdef f(\x_t, t) + \frac{1}{2}g(t)^2 \nabla q_t(\x_t).
\end{equation}
If \(\x_0 \sim p_{\x}\), then the trajectory \(\x_t\) satisfies \(\x_t \sim p_t(\x_t)\) for all \(t \in [0,T]\). We can compute the potential of an image
\(x_0\) with the instantaneous change of variables formula~\cite{chen2018neural}:
\begin{equation}\label{eq:likelihood_computation}
    -\log p_{\x}(x_0) = -\log p_T(x_T) - \int_{0}^T \nabla \cdot v(x_t, t) \,dt,
\end{equation}
where \(x_t\) is the trajectory of the ODE in \cref{eq:ode_flow}. % can be obtained by solving the ODE \eqref{eq:ode_flow}.
The divergence term \(\nabla \cdot v(x_t, t)\) can be approximated by the Skilling-Hutchinson trace estimator~\cite{Skilling1989}:
\begin{equation*}
    \nabla \cdot v(x_t, t) = \mathbb{E}_{\boldsymbol{u}} \left[\boldsymbol{u}^T  J_v (x_t, t) \boldsymbol{u}  \right],
\end{equation*}
where \(J_v\) is the Jacobian of \(v(\cdot, t)\), and \(\boldsymbol{u}\) is a random variable with zero mean and identity covariance. In our
implementation, we use a Rademacher random variable.
The vector-Jacobian product \(\boldsymbol{u}^T J_v (x_t, t)\) is computed using reverse-mode automatic differentiation with roughly the same cost as
evaluating \(v(x_t, t)\).
The Skilling-Hutchinson trace estimator is unbiased so by averaging multiple runs, we can approximate the divergence.
The integral can be evaluated using numerical ODE solver.
Finally, at time \(T\), the random variable \(\x_T\) follows a Gaussian distribution with zero mean and standard deviation \(\sigma(T)\) (depending on the
model and on the dataset), which allows us to compute the term \(\log p_T(x_T)\) in \cref{eq:likelihood_computation}.

In all of our experiments, we use the Runge-Kutta 45 numerical scheme, provided by \texttt{scipy.integrate.solve\_ivp} with tolerance value
\texttt{atol=1e-5} and \texttt{rtol=1e-5}.
We use a similar implementation to~\cite{song2021scorebased}, but we compute the exact value of the potential instead of the bit-ber-dim value in the
original implementation.
This process is numerically intensive, as it requires hundreds of evaluations of the neural network \(\nabla q_t\).

\paragraph{Datasets, pretrained models and implementation details}
We use a model\footnote{Pre-trained model: \href{https://github.com/yang-song/score_sde_pytorch}{https://github.com/yang-song/score\_sde\_pytorch}}
trained on the \(256 \times 256\) FFHQ dataset from Song \etal~\cite{song2021scorebased} and three models\footnote{Pre-trained model:
\href{https://github.com/NVlabs/edm}{https://github.com/NVlabs/edm}} trained on the ImageNet, FFHQ and AFHQ datasets at a \(64 \times 64\) resolution
from the widely used EDM framework by Karras \etal~\cite{Karras2022edm}.
The latter are more recent and produce higher quality images.
For brevity, we refer to these models throughout the paper as FFHQ-256, ImageNet-64, FFHQ-64, and AFHQ-64 respectively.
ImageNet-64 is a class-conditional model, throughout the paper, we use the first 10 classes of the ImageNet dataset.
We summarize the pre-trained models used in this paper in \cref{tab:supp_pretrained_models}.
\begin{table}[htbp]
    \centering
    \caption{Pre-trained diffusion models used in our study. The computation time was evaluated using a single NVIDIA V100 SMX2-16GB GPU and with 10
    CPU cores from an  Intel Cascade Lake 6248 processor.}
    \label{tab:supp_pretrained_models}
    \resizebox{\linewidth}{!}{
        \begin{tabular}{lcccc}
            \hline %   \toprule
            Pre-trained model                             & FFHQ-256~\cite{song2021scorebased} & ImageNet-64~\cite{Karras2022edm} &
            FFHQ-64~\cite{Karras2022edm} &  AFHQ-64~\cite{Karras2022edm} \\
            \hline %   \midrule
            Train dataset             & FFHQ & ImageNet & FFHQ & AFQH \\
            SDE/ODE formulation         & VE SDE & EDM & EDM  & EDM \\
            Resolution & \(256 \times 256 \times 3\) & \(64 \times 64 \times 3\) & \(64 \times 64 \times 3\) & \(64 \times 64 \times 3\)\\
            Num. parameters (million)    &    \(66\)         &    \(296\)             &         \(63\)    &     \(63\)         \\
            Model eval. (sec./image)         & \(0.036\)    & \(0.014\)  & \(0.006\)     & \(0.006\)     \\
            Potential eval. (sec/image) & \(8.461\)    & \(10.26\)  & \(5.05\)    & \(5.05\)\\
            \hline %   \bottomrule
        \end{tabular}
    }
\end{table}

% ------------------------------------------------------------

\null
\newpage

\section{Additional numerical result}
We provide additional numerical results to complement those in \cref{sec:general_kernel}.

\begin{figure}[htbp]
    \def\imgwidth{0.15\linewidth}
    \centering
    \raisebox{0.08\linewidth}{\makebox[0.25cm][c]{\(\bar x\)}}
    \includegraphics[width=\imgwidth]{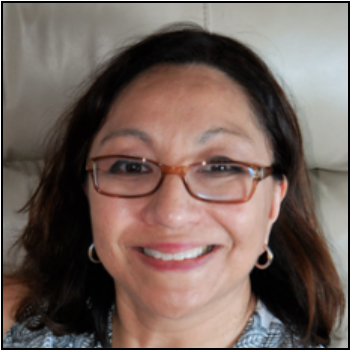}
    \includegraphics[width=\imgwidth]{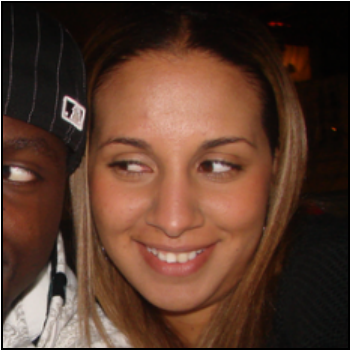}
    \includegraphics[width=\imgwidth]{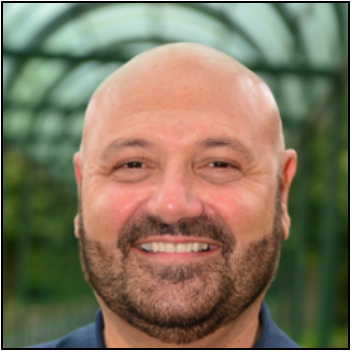}
    \includegraphics[width=\imgwidth]{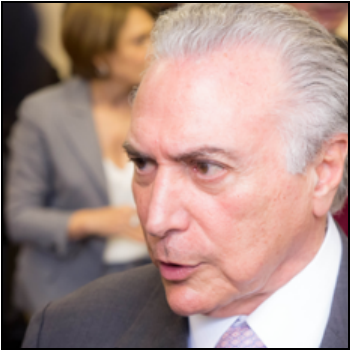}
    \includegraphics[width=\imgwidth]{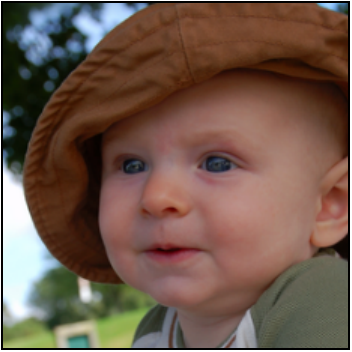}
    \\
    \raisebox{0.08\linewidth}{\makebox[0.25cm][c]{\(\bar h\)}}
    \includegraphics[width=\imgwidth]{./Images/gretsi/uniform_motion_45_kernel.pdf}
    \includegraphics[width=\imgwidth]{./Images/gretsi/uniform_levin_kernel.pdf}
    \includegraphics[width=\imgwidth]{./Images/gretsi/uniform_diffraction_simple_kernel.pdf}
    \includegraphics[width=\imgwidth]{./Images/gretsi/uniform_diffraction_complex_kernel.pdf}
    \includegraphics[width=\imgwidth]{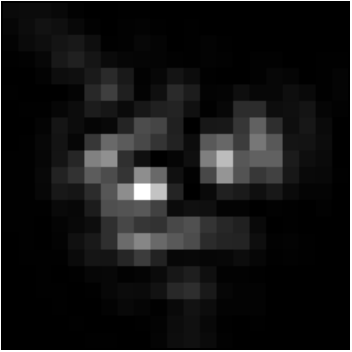}
    \\
    \raisebox{0.08\linewidth}{\makebox[0.25cm][c]{\(y\)}}
    \includegraphics[width=\imgwidth]{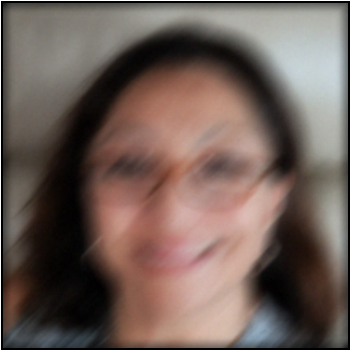}
    \includegraphics[width=\imgwidth]{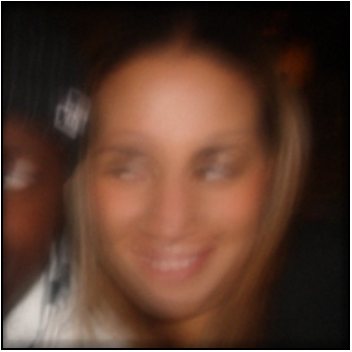}
    \includegraphics[width=\imgwidth]{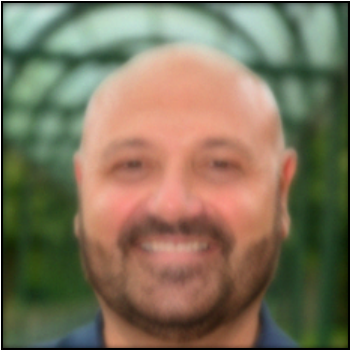}
    \includegraphics[width=\imgwidth]{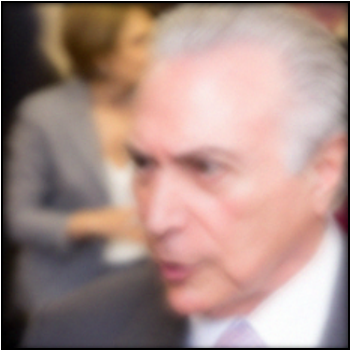}
    \includegraphics[width=\imgwidth]{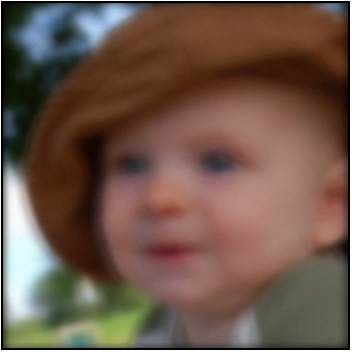}
    \\
    \raisebox{0.08\linewidth}{\makebox[0.25cm][c]{\(\hat x\)}}
    \includegraphics[width=\imgwidth]{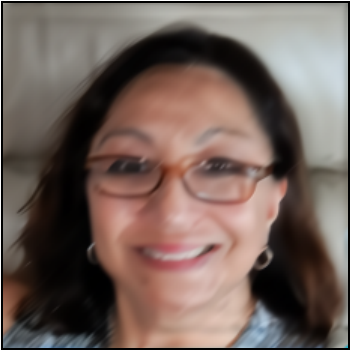}
    \includegraphics[width=\imgwidth]{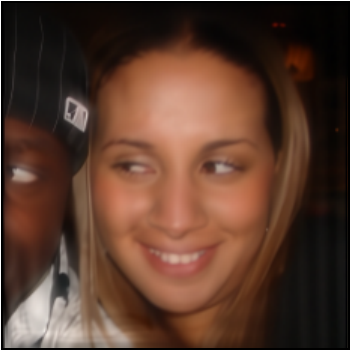}
    \includegraphics[width=\imgwidth]{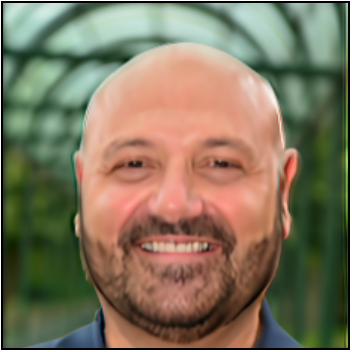}
    \includegraphics[width=\imgwidth]{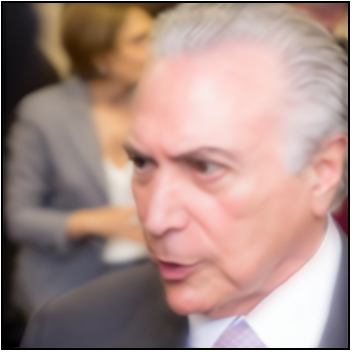}
    \includegraphics[width=\imgwidth]{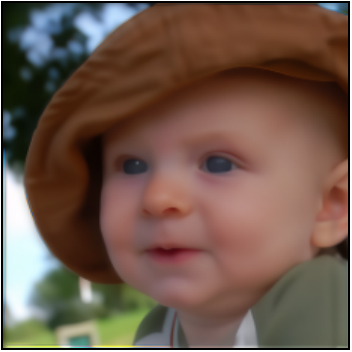}
    \\
    \raisebox{0.08\linewidth}{\makebox[0.25cm][c]{\(\hat h\)}}
    \includegraphics[width=\imgwidth]{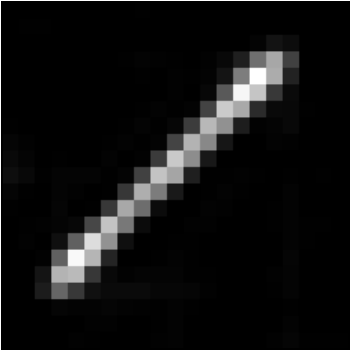}
    \includegraphics[width=\imgwidth]{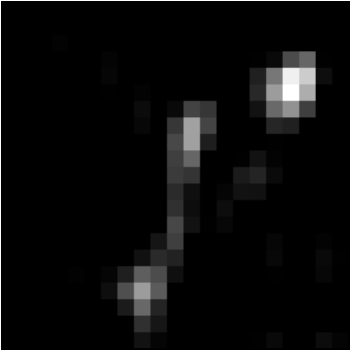}
    \includegraphics[width=\imgwidth]{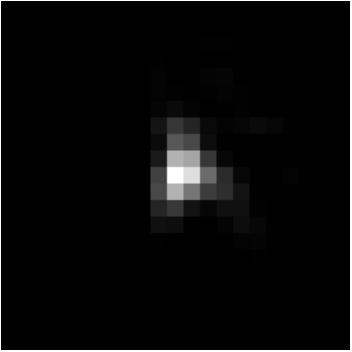}
    \includegraphics[width=\imgwidth]{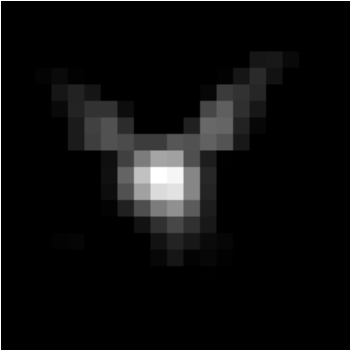}
    \includegraphics[width=\imgwidth]{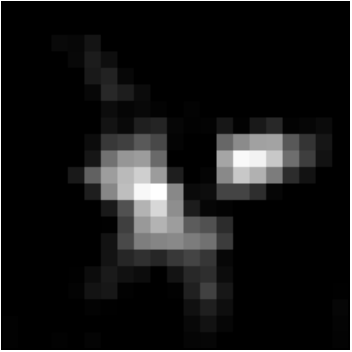}
    \\
    \caption{More numerical results with a simplex constraint using different pairs \((\bar{x}, \bar{h})\), similarly as
        \cref{fig:numerical_result_simulated}. The algorithm \cref{algo:main_algorithm} is initialized by a uniform kernel. The noise level is \(\sigma =
    0.01\). The kernel's shape is recovered approximately and the estimated images look sharper.}
    \label{fig:numerical_result_gretsi_endpoints}
\end{figure}

\begin{figure}[htbp]
    \centering
    \def\half{0.4841\linewidth}
    \def\inside{0.935\linewidth}
    \def\labelwidth{0.05\linewidth}
    \subfloat[Noise level \(\sigma = 0.02\)]{
        \begin{minipage}{\half}
            \centering
            \begin{minipage}{\labelwidth}
                \rotatebox{90}{\footnotesize \(\bar{x}\)}
            \end{minipage}
            \begin{minipage}{\inside}
                \centering
                \includegraphics[width=\linewidth]{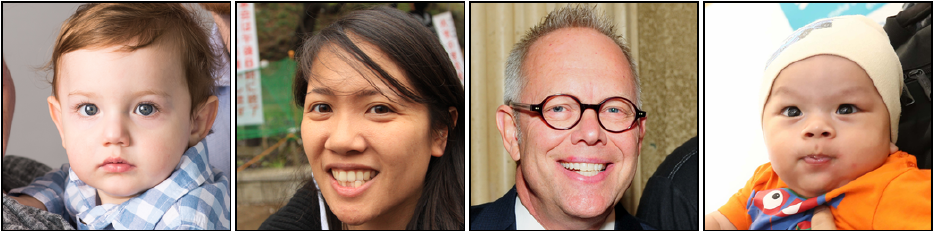}
            \end{minipage}
            \begin{minipage}{\labelwidth}
                \rotatebox{90}{\footnotesize\(y\)}
            \end{minipage}
            \begin{minipage}{\inside}
                \centering
                \includegraphics[width=\linewidth]{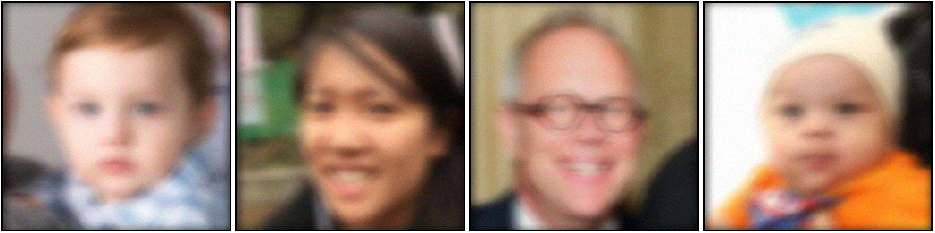}
            \end{minipage}
            \begin{minipage}{\labelwidth}
                \rotatebox{90}{\footnotesize \(\hat{x}_1\) and \(\hat{h}_1\)}
            \end{minipage}
            \begin{minipage}{\inside}
                \centering
                \includegraphics[width=\linewidth]{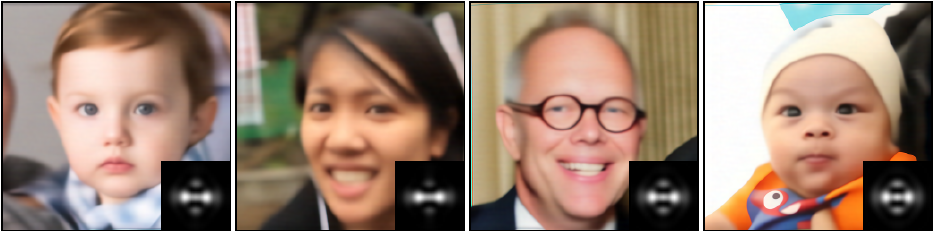}
            \end{minipage}
            \begin{minipage}{\labelwidth}
                \rotatebox{90}{\footnotesize \(\hat{x}_2\) and \(\hat{h}_2\)}
            \end{minipage}
            \begin{minipage}{\inside}
                \centering
                \includegraphics[width=\linewidth]{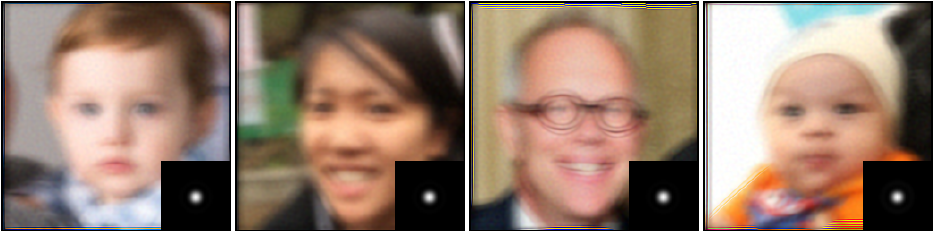}
            \end{minipage}
            % -------------------------
            \begin{minipage}{0.05\linewidth}
                \rotatebox{90}{\footnotesize}
            \end{minipage}
            \begin{minipage}{0.3\linewidth}
                \centering
                \includegraphics[width=\linewidth]{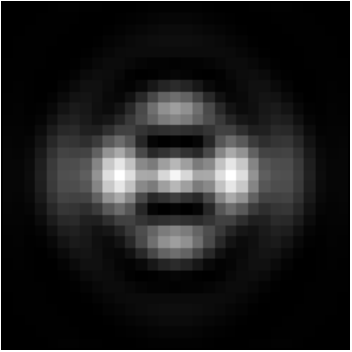}
            \end{minipage}
            \begin{minipage}{0.3\linewidth}
                \centering
                \includegraphics[width=\linewidth]{./Images/zernike/radius_0.15_natural_larger_3_sigma_0.02_adam_zernike_iter_400_kernel_bar.pdf}
            \end{minipage}
            \begin{minipage}{0.3\linewidth}
                \centering
                \includegraphics[width=\linewidth]{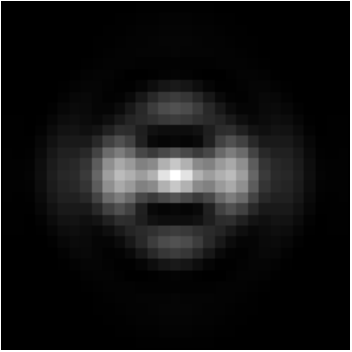}
            \end{minipage}
            \begin{minipage}{\labelwidth}

            \end{minipage}
            \begin{minipage}{0.3\linewidth}
                \centering
                \vspace{0.1cm}
                \footnotesize \(\bar{h}\)
            \end{minipage}
            \begin{minipage}{0.3\linewidth}
                \centering
                \vspace{0.1cm}
                \footnotesize \(h_{\mathrm{init. 1}}\)
            \end{minipage}
            \begin{minipage}{0.3\linewidth}
                \centering
                \vspace{0.1cm}
                \footnotesize \(h_{\mathrm{init. 2}}\)
            \end{minipage}

        \end{minipage}
    }
    \subfloat[Noise level \(\sigma = 0.01\)]{
        \begin{minipage}{\half}
            \begin{minipage}{\inside}
                \centering
                \includegraphics[width=\linewidth]{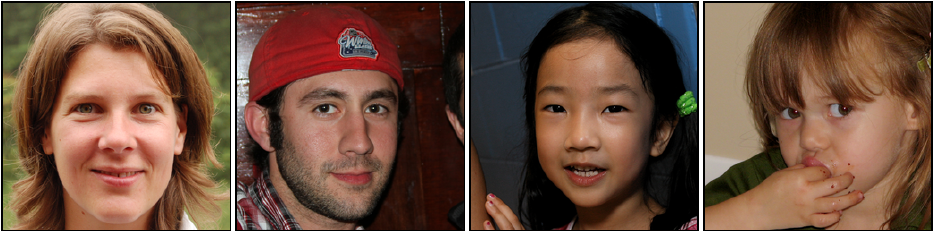}
            \end{minipage}
            \begin{minipage}{\inside}
                \centering
                \includegraphics[width=\linewidth]{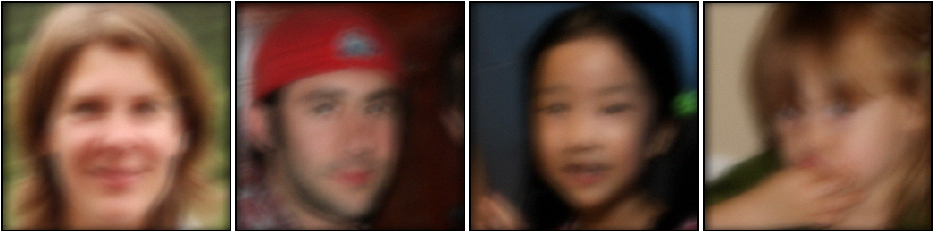}
            \end{minipage}
            \begin{minipage}{\inside}
                \centering
                \includegraphics[width=\linewidth]{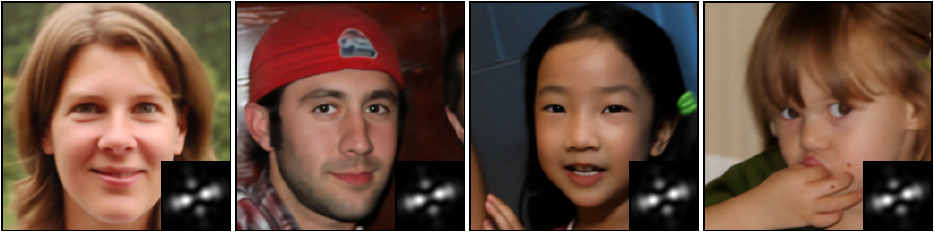}
            \end{minipage}
            \begin{minipage}{\inside}
                \centering
                \includegraphics[width=\linewidth]{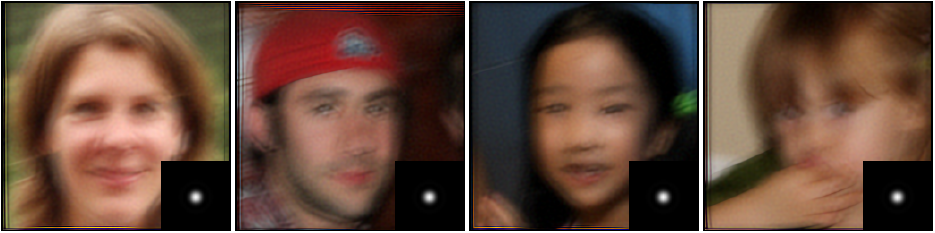}
            \end{minipage}
            % -------------------------
            % \begin{minipage}{\labelwidth}
            %     \rotatebox{90}{\footnotesize}
            % \end{minipage}
            \begin{minipage}{0.3\linewidth}
                \centering
                \includegraphics[width=\linewidth]{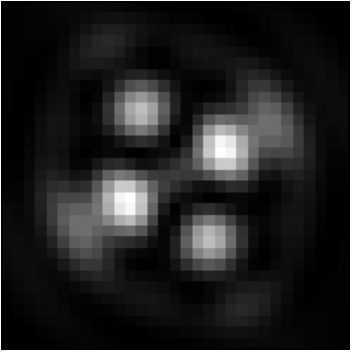}
            \end{minipage}
            \begin{minipage}{0.3\linewidth}
                \centering
                \includegraphics[width=\linewidth]{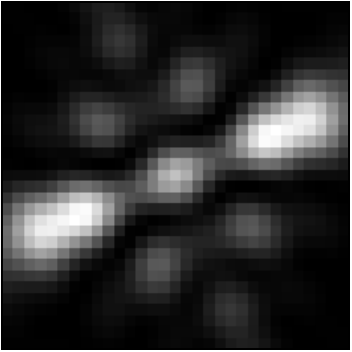}
            \end{minipage}
            \begin{minipage}{0.3\linewidth}
                \centering
                \includegraphics[width=\linewidth]{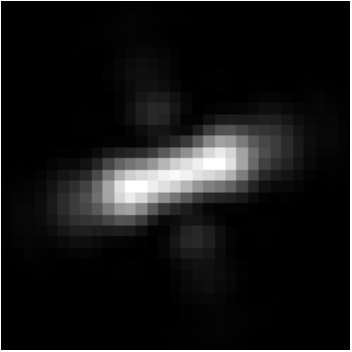}
            \end{minipage}
            \begin{minipage}{0.05\linewidth}
                \rotatebox{90}{\footnotesize}
            \end{minipage}
            \begin{minipage}{0.3\linewidth}
                \centering
                \vspace{0.1cm}
                \footnotesize \(\bar{h}\)
            \end{minipage}
            \begin{minipage}{0.3\linewidth}
                \centering
                \vspace{0.1cm}
                \footnotesize \(h_{\mathrm{init. 1}}\)
            \end{minipage}
            \begin{minipage}{0.3\linewidth}
                \centering
                \vspace{0.1cm}
                \footnotesize \(h_{\mathrm{init. 2}}\)
            \end{minipage}
            \begin{minipage}{0.05\linewidth}
                \rotatebox{90}{\footnotesize}
            \end{minipage}
        \end{minipage}
    }
    \caption{More examples on the influence of initialization for blind deconvolution, with diffraction-limited blur kernels as in
        \Cref{fig:zernike_joint_min_larger_kernel}. The initialization are \(h_{\mathrm{init. 1}} = h(\bar{\theta} + 0.15 \cdot
        \mathrm{sign}(\bar{\theta}))\) and \(h_{\mathrm{init. 2}} = h(\bar{\theta} - 0.075 \cdot \mathrm{sign}(\bar{\theta}))\) and \(x_0 = y\). Again, using
    large kernels for initialization leads to a decent kernel reconstruction, while small kernels lead to the no-blur solution.}
    \label{fig:supp_more_zernike}
\end{figure}

% ------------------------------------------------------------
\null
\pagebreak
\null
\newpage
\section{Proofs}
We start by the following lemma which characterizes the positiveness of a block matrix.
\begin{lemma}
    \label{lem:invertibilityHessian}
    Let \(A \in \R^{M \times N}\), \(J \in \R^{M \times P}\) and \(D \in \R^{N \times N}\) with \(D\) symmetric, positive semi-definite. Consider the
    block matrix:
    \begin{align*}
        H =
        \begin{pmatrix}
            A^T A + D & A^T J \\
            J^T A & J^T J
        \end{pmatrix}.
    \end{align*}
    Then \(H\) is positive definite if and only if:
    \begin{enumerate}
        \item \(\ker J = \{0\}\),
        \item \(A \ker D \cap \mathrm{Im}\ J = \{0\}\),
        \item \(\ker D \cap \ker A = \{0\}\).
    \end{enumerate}
\end{lemma}

\begin{proof}
    Since \(D\) is symmetric, positive semi-definite, let \(\sqrt{D}\) be its symmetric semi-definite square root. Define the matrix \(L\) as:
    \begin{align*}
        L =
        \begin{pmatrix}
            A & J \\
            \sqrt{D} & 0
        \end{pmatrix}.
    \end{align*}
    We observe that \(H = L^T L\). Thus, \(H\) is positive semi-definite, and it is positive definite if and only if \(\ker L = \{0\}\).

    A vector \((u,v) \in \R^N \times \R^P\) is in \(\ker L\) if and only if:
    \begin{equation} \label{eq:kerSystem}
        \begin{cases}
            Au + Jv = 0 \\
            \sqrt{D}u = 0
        \end{cases}
        \iff
        \begin{cases}
            Au + Jv = 0 \\
            u \in \ker D
        \end{cases}
    \end{equation}

    \paragraph{Sufficiency:} Assume the three conditions hold. Let \((u,v) \in \ker L\). From \eqref{eq:kerSystem}, \(u \in \ker D\) and \(Au = -Jv\). Thus
    \(Au \in A \ker D \cap \mathrm{Im}\ J\). By condition (2), this implies \(Au = 0\). Since \(u \in \ker D\) and \(Au=0\), condition (3)
    implies \(u=0\). The
    system then reduces to \(Jv=0\), which by condition (1) implies \(v=0\). Thus \(\ker L = \{0\}\).

    \paragraph{Necessity:} Assume that \(H\) is positive definite, so that \(\ker L = \{0\}\).
    \begin{itemize}
        \item If \(\ker J \neq \{0\}\), there exists \(v \neq 0\) such that \(Jv=0\). Then \((0,v) \in \ker L\).
        \item If \(y \in A \ker D \cap \mathrm{Im}\ J\) is non-zero, there exist \(u \in \ker D\) and \(v\) such that \(Au = y = -Jv\). Then \(Au + Jv = 0\)
            and \(u \in \ker D\), so \((u,v) \in \ker L\). Note that \(u \neq 0\) because \(Au = y \neq 0\).
        \item If \(u \in \ker D \cap \ker A\) with \(u \neq 0\), then \((u, 0) \in \ker L\).
    \end{itemize}
    In all cases, violation of a condition implies \(\ker L \neq \{0\}\), which is a contradiction.
\end{proof}

% ------------------------------------------------
% ------------------------------------------------
\subsection{Proof of \cref{thm:stability_nonblind}}
We now prove the stability of the non-blind MAP estimator, as stated in \cref{thm:stability_nonblind}.
\begin{proof}
    Define
    \[ F(x,y) \eqdef \nabla_x \Lc(x;y) = \frac{1}{\sigma^2} A^T(Ax-y) + \nabla q(x).\]
    % \paragraph{Step 1: Stationarity at the ground truth}
    Since \(\bar y = A\bar x\) and \(\bar x\) lies on the critical submanifold
    \(\mathcal M\) of \(q\), we have
    \[ \nabla q(\bar x)=0,\]
    hence
    \[ F(\bar x,\bar y) = \frac{1}{\sigma^2} A^T(A\bar x-\bar y) + \nabla q(\bar x) = 0.\]
    % \paragraph{Step 2: Positivity of the Hessian}
    The differential of \(F\) with respect to \(x\) at \((\bar x,\bar y)\) is
    \[ \bar H \eqdef \partial_x F(\bar x,\bar y) =\frac{1}{\sigma^2}  A^T A + \nabla^2 q(\bar x).\]
    Since \(q\) is locally Morse--Bott at \(\bar x\) with critical submanifold
    \(\mathcal M\), we have
    \[
        \ker \nabla^2 q(\bar x)=T_{\bar x}\mathcal M.
    \]
    Moreover, since \(\nabla^2 q(\bar x) \succeq 0\), applying \cref{lem:invertibilityHessian} with \(D = \nabla^2 q(\bar x)\) and \(J=0\), we deduce that
    \(\bar H\) is positive definite:
    \[\mu \eqdef \lambda_{\min}(\bar H) >0,\]
    where  \(\mu\) as the smallest eigenvalue of \(\bar H\).
    % \paragraph{Step 3: Application of the implicit function theorem}
    Since \(\bar H\) is invertible, the implicit function theorem yields
    neighborhoods \(U\) of \(\bar x\) and \(V\) of \(\bar y\), and a \(\mathcal{C}^1\) map
    \(y\mapsto x(y)\) such that
    \[ F(x(y),y)=0, \qquad x(\bar y)=\bar x, \qquad \nabla^2_x L(x(y),y) \succ 0. \]
    % \paragraph{Step 4: Linearization}
    The last inequality is a consequence of the continuity of \(\nabla^2_x L\) and the fact that \(\nabla^2_x L(\bar x, \bar y) = \bar H \succ 0\).

    The first equality reads \(\frac{1}{\sigma^2} A^T(Ax(y)-y) + \nabla q(x(y))=0.\)
    Differentiating w.r.t. \(y\) yields
    \[ \left( \frac{1}{\sigma^2} A^T A  + \nabla^2 q(x(y)) \right) \mathrm{Jac}_x (y) - \frac{1}{\sigma^2} A^T = 0, \qquad \forall y \in V.\]
    Taking \(y = \bar y\), we have
    \[ \bar H \mathrm{Jac}_x(\bar y) - \frac{1}{\sigma^2} A^T = 0, \qquad \text{hence} \qquad \mathrm{Jac}_x(\bar y) = \frac{1}{\sigma^2} \bar H^{-1} A^T.\]
    Therefore, first-order approximation yields
    \[x(y) = x(\bar y) + \mathrm{Jac}_x(\bar y) (y - \bar y)  + o(\|y-\bar y\|)\]
    or equivalently
    \[ x(y) - \bar x = \frac{1}{\sigma^2} \bar H^{-1} A^T (y - \bar y) + o(\|y-\bar y\|).\]
    % \paragraph{Step 5: Stability bound}
    Since \(\|\bar H^{-1}\|=1/\lambda_{\min}(\bar H)=1/\mu\), we obtain the stability bound as
    \[\|x(y)-\bar x\| \le \frac{\|A\|}{\mu \sigma^2} \|y-\bar y\| + o(\|y-\bar y\|).\]
\end{proof}

\subsection{Proof of \cref{th:stabilityLocalMinima} and \cref{th:stability_bounds}}
\label{sec:proof_stability_local_minima}
We provide a joint proof of \cref{th:stabilityLocalMinima} and \cref{th:stability_bounds}, as they are closely related.
\begin{proof}
    The proof proceeds as follows
    \paragraph{First-order optimality}
    Define the gradient of \(L\) as
    \[F(x,\theta,y) \eqdef \nabla_{x,\theta}\Lc(x,\theta;y).\]
    A direct computation gives
    \[F(x,\theta,y) =
        \begin{pmatrix}
            \frac{1}{\sigma^2} A(\theta)^T (A(\theta)x-y)+\nabla q(x) \\
            \frac{1}{\sigma^2} J_x(\theta)^T(A(\theta)x-y)
        \end{pmatrix},
    \]
    where \(J_x(\theta) = \frac{\partial}{\partial \theta} \left( A(\theta) x \right) \in \R^{M \times P}\) denotes the Jacobian of the measurements
    w.r.t the parameters \(\theta\).
    Since \(A(\bar\theta)\bar x=\bar y\) and \(\nabla q(\bar x)=0\),
    we obtain
    \[F(\bar x,\bar\theta,\bar y)=0.\]
    \paragraph{Expression of the joint Hessian}
    Differentiating \(F\) with respect to \((x,\theta)\) at \((\bar x,\bar\theta,\bar y)\) gives the joint Hessian
    \[\bar H \eqdef \nabla^2_{x,\theta}\Lc(\bar x,\bar\theta;\bar y)
        =
        \begin{pmatrix}
            \bar H_{xx} & \bar H_{x\theta} \\
            \bar H_{\theta x} & \bar H_{\theta\theta}
        \end{pmatrix}
        =  \frac{1}{\sigma^2}
        \begin{pmatrix}
            A(\bar\theta)^T A(\bar\theta) + \sigma^2 \nabla^2 q(\bar x) & A(\bar \theta)^T J_{\bar{x}}(\bar \theta)\\
            J_{\bar x}(\bar\theta)^T A(\bar\theta) & J_{\bar x}(\bar\theta)^T J_{\bar x}(\bar\theta)
        \end{pmatrix}.
    \]

    Under the three geometric compatibility conditions and \(\nabla^2 q(\bar x) \succeq 0\), \cref{lem:invertibilityHessian} implies that \(\bar H\) is
    positive definite:
    \[ \bar H \succeq \mu I_{N+P}, \qquad \mu \eqdef \lambda_{\min}(\bar H)>0 \text{ is the smallest eigenvalue}.\]
    % -----------------------------------------------------
    \paragraph{Application of the implicit function theorem}
    Since \(F\) is \(\mathcal{C}^1\) and \(D_{(x,\theta)}F(\bar x,\bar\theta,\bar y)=\bar H\)
    is invertible, the implicit function theorem yields neighborhoods
    \(U\) of \((\bar x,\bar\theta)\) and \(V\) of \(\bar y\) and a unique \(\mathcal{C}^1\) map \(y \mapsto (x(y),\theta(y))\) such that
    \[F(x(y),\theta(y),y)=0,  \qquad (x(\bar y), \theta(\bar y)) = (\bar x, \bar \theta), \qquad \nabla^2_{x, \theta} L(x(y),\theta(y);y) \succ 0.\]
    The last inequality is a consequence of the continuity of \(\nabla^2_{x,\theta} L\) and  \(\nabla^2_{x,\theta} L(\bar x, \bar \theta;\bar y) \succ 0\).

    \paragraph{First-order expansion}
    The first equality reads
    \begin{equation}
        \frac{1}{\sigma^2}
        \begin{pmatrix}
            A(\theta(y))^T (A(\theta(y)) x(y) - y) + \sigma^2 \nabla q(x(y)) \\
            J_x(\theta(y))^T(A(\theta(y))x(y) - y)
        \end{pmatrix} = 0.
    \end{equation}
    Differentiating the above equality w.r.t \(y\) then evaluating at \((\bar x, \bar \theta)\), we have
    \begin{equation}
        \begin{pmatrix}
            \bar H_{xx} & \bar H_{x\theta} \\
            \bar H_{\theta x} & \bar H_{\theta\theta}
        \end{pmatrix}
        \begin{pmatrix}
            \mathrm{Jac}_x(\bar y) \\
            \mathrm{Jac}_\theta(\bar y)
        \end{pmatrix}
        =
        \frac{1}{\sigma^2}
        \begin{pmatrix}
            A(\bar\theta)^T \\
            J_{\bar x}(\bar\theta)^T
        \end{pmatrix},
    \end{equation}
    where \(\mathrm{Jac}_x\) and \(\mathrm{Jac}_\theta\) denote the Jacobian matrices of \(x(y)\) and \(\theta(y)\) respectively.
    Therefore,
    \begin{equation}
        \begin{pmatrix}
            \mathrm{Jac}_x(\bar y) \\
            \mathrm{Jac}_\theta(\bar y)
        \end{pmatrix}
        = \frac{1}{\sigma^2} \bar{H}^{-1}
        \begin{pmatrix}
            A(\bar\theta)^T \\
            J_{\bar x}(\bar\theta)^T
        \end{pmatrix}.
    \end{equation}
    Using the Schur complement formula for the inverse of a block matrix, we have
    \begin{equation}\label{eq:analytical_joint_hessian}
        \bar H^{-1} =
        \begin{pmatrix}
            S_{x}^{-1}        & -S_x^{-1} \bar H_{x \theta} \bar H_{\theta \theta}^{-1} \\
            -\bar H_{\theta \theta}^{-1}\bar H_{\theta x} S_x^{-1} &  \bar H_{\theta \theta}^{-1} + \bar H_{\theta \theta}^{-1}\bar H_{\theta x}
            S_x^{-1} \bar H_{x \theta} \bar H_{\theta \theta}^{-1}
        \end{pmatrix},
    \end{equation}
    where \(S_x = \bar H_{xx} - \bar H_{x\theta} \bar H_{\theta\theta}^{-1} \bar H_{\theta x}\) is the Schur complement of \(\bar H\) with respect to
    \(\bar H_{\theta\theta}\).
    We deduce that
    \begin{align}
        \mathrm{Jac}_x(\bar y) &= \frac{1}{\sigma^2}  S_x^{-1} \left(A(\bar\theta)^T-\bar H_{x \theta} \bar H_{\theta \theta}^{-1} J_{\bar
        x}(\bar\theta)^T \right),  \\
        \mathrm{Jac}_\theta(\bar y) &= -\bar H_{\theta \theta}^{-1} \bar H_{\theta x} S_x^{-1} \frac{1}{\sigma^2} A(\bar\theta)^T + \left( \bar
        H_{\theta \theta}^{-1} + \bar H_{\theta \theta}^{-1}\bar H_{\theta x} S_x^{-1} \bar H_{x \theta} \bar H_{\theta \theta}^{-1} \right)
        \frac{1}{\sigma^2} J_{\bar x}(\bar\theta)^T  \notag \\
        &= - \frac{1}{\sigma^2}  \bar H_{\theta \theta}^{-1} \bar H_{\theta x} S_x^{-1} \left( A(\bar\theta)^T -  \bar H_{x \theta} \bar H_{\theta
        \theta}^{-1} J_{\bar x}(\bar\theta)^T  \right) + \frac{1}{\sigma^2} \bar H_{\theta \theta}^{-1} J_{\bar x}(\bar\theta)^T
    \end{align}
\end{proof}
\subsection{Proof of \cref{theorem:no_blur_solution}}\label{subsec:proof_no_blur_solution}

\begin{proof}
    Let \((\hat x, \hat h)\) be a minimizer of \cref{eq:def_ly}. If \(\hat h \neq \delta \), we can replace \((\hat x, \hat h)\) with \((\hat h \conv \hat
    x, \delta)\) and get a lower cost function. Indeed, the point \((\hat h \conv \hat x, \delta)\) does not modify the value of the data fidelity term
    in \cref{eq:def_ly} and since we assumed that \(q(\hat x) \geq q(\hat h \conv \hat x)\), \((\hat h \conv \hat x_0, \delta)\) is also a solution.
    This shows that the Dirac kernel is a solution to the joint MAP problem.
    If the inequality were strict in \cref{eq:keyinequality}, the solution would  be unique and satisfy \(\hmap=\delta\).
    Note that when \(\hmap\) is the Dirac mass, the estimated image coincides with a MAP-denoised image:
    \begin{equation}
        \xmap(y) = \argmin_{x} \frac{1}{2\sigma^2} \norm{x - y}_2^2 + q(x).
    \end{equation}
\end{proof}
\bibliographystyle{siamplain}
\bibliography{references}

@String(CVPR= {IEEE Conf. Comput. Vis. Pattern Recog.})

@String(CVPR  = {CVPR})

@inproceedings{bora2017compressed,
  title={Compressed sensing using generative models},
  author={Bora, Ashish and Jalal, Ajil and Price, Eric and Dimakis, Alexandros G},
  booktitle={International conference on machine learning},
  pages={537--546},
  year={2017},
  organization={PMLR}
}

@article{kech2017optimal,
  title={Optimal injectivity conditions for bilinear inverse problems with applications to identifiability of deconvolution problems},
  author={Kech, Michael and Krahmer, Felix},
  journal={SIAM Journal on Applied Algebra and Geometry},
  volume={1},
  number={1},
  pages={20--37},
  year={2017},
  publisher={SIAM}
}

@article{ahmed2013blind,
  title={Blind deconvolution using convex programming},
  author={Ahmed, Ali and Recht, Benjamin and Romberg, Justin},
  journal={IEEE Transactions on Information Theory},
  volume={60},
  number={3},
  pages={1711--1732},
  year={2013},
  publisher={IEEE}
}

@inproceedings{popeintrinsic,
  title={The Intrinsic Dimension of Images and Its Impact on Learning},
  author={Pope, Phil and Zhu, Chen and Abdelkader, Ahmed and Goldblum, Micah and Goldstein, Tom},
  booktitle={International Conference on Learning Representations}
}

@inproceedings{lee2016gradient,
  title={Gradient descent only converges to minimizers},
  author={Lee, Jason D and Simchowitz, Max and Jordan, Michael I and Recht, Benjamin},
  booktitle={Conference on learning theory},
  pages={1246--1257},
  year={2016},
  organization={PMLR}
}

@inproceedings{Karras2022edm,
  author    = {Tero Karras and Miika Aittala and Timo Aila and Samuli Laine},
  title     = {Elucidating the Design Space of Diffusion-Based Generative Models},
  booktitle = {Proc. NeurIPS},
  year      = {2022}
}

@article{pidstrigach2022score,
  title={Score-based generative models detect manifolds},
  author={Pidstrigach, Jakiw},
  journal={Advances in Neural Information Processing Systems},
  volume={35},
  pages={35852--35865},
  year={2022}
}

@article{rebjock2024fast,
  title={Fast convergence to non-isolated minima: four equivalent conditions for c 2 functions},
  author={Rebjock, Quentin and Boumal, Nicolas},
  journal={Mathematical Programming},
  pages={1--49},
  year={2024},
  publisher={Springer}
}

@incollection{lee2003smooth,
  title={Smooth manifolds},
  author={Lee, John M},
  booktitle={Introduction to smooth manifolds},
  pages={1--29},
  year={2003},
  publisher={Springer}
}

@article{ANDERSON1982313,
title = {Reverse-time diffusion equation models},
journal = {Stochastic Processes and their Applications},
volume = {12},
number = {3},
pages = {313-326},
year = {1982},
issn = {0304-4149},
doi = {https://doi.org/10.1016/0304-4149(82)90051-5},
url = {https://www.sciencedirect.com/science/article/pii/0304414982900515},
author = {Brian D.O. Anderson}
}

@article{reehorst2018regularization,
  title={Regularization by denoising: Clarifications and new interpretations},
  author={Reehorst, Edward T and Schniter, Philip},
  journal={IEEE transactions on computational imaging},
  volume={5},
  number={1},
  pages={52--67},
  year={2018},
  publisher={IEEE}
}

@INPROCEEDINGS{levin2009understanding,
  author={Levin, Anat and Weiss, Yair and Durand, Fredo and Freeman, William T.},
  booktitle={2009 IEEE Conference on Computer Vision and Pattern Recognition}, 
  title={Understanding and evaluating blind deconvolution algorithms}, 
  year={2009},
  volume={},
  number={},
  pages={1964-1971},
  doi={10.1109/CVPR.2009.5206815}}

@inproceedings{hurault2022gradient,
  title={Gradient Step Denoiser for convergent Plug-and-Play},
  author={Hurault, Samuel and Leclaire, Arthur and Papadakis, Nicolas},
  booktitle={International Conference on Learning Representations},
  year={2021}
}

@inproceedings{song2021scorebased,
  title={Score-Based Generative Modeling through Stochastic Differential Equations},
  author={Song, Yang and Sohl-Dickstein, Jascha and Kingma, Diederik P and Kumar, Abhishek and Ermon, Stefano and Poole, Ben},
  booktitle={International Conference on Learning Representations},
  year={2020}
}

@book{nesterov2013introductory,
  title={Introductory lectures on convex optimization: A basic course},
  author={Nesterov, Yurii},
  volume={87},
  year={2013},
  publisher={Springer Science \& Business Media}
}

@inproceedings{pan2016blind,
  title={Blind image deblurring using dark channel prior},
  author={Pan, Jinshan and Sun, Deqing and Pfister, Hanspeter and Yang, Ming-Hsuan},
  booktitle={Proceedings of the IEEE conference on computer vision and pattern recognition},
  pages={1628--1636},
  year={2016}
}

@article{chan1998total,
  author =        {Chan, Tony F and Wong, Chiu-Kwong},
  journal =       {IEEE transactions on Image Processing},
  number =        {3},
  pages =         {370--375},
  publisher =     {IEEE},
  title={         {Total variation blind deconvolution}},
  volume =        {7},
  year =          {1998},
}

@inproceedings{pan2014deblurring,
  title={Deblurring text images via L0-regularized intensity and gradient prior},
  author={Pan, Jinshan and Hu, Zhe and Su, Zhixun and Yang, Ming-Hsuan},
  booktitle={Proceedings of the IEEE Conference on Computer Vision and Pattern Recognition},
  pages={2901--2908},
  year={2014}
}

@article{pustelnik2016wavelet,
  title={Wavelet-based image deconvolution and reconstruction},
  author={Pustelnik, Nelly and Benazza-Benhayia, Amel and Zheng, Yuling and Pesquet, Jean-Christophe},
  journal={Wiley encyclopedia of electrical and electronics engineering},
  year={2016}
}

@article{shajkofci2020spatially,
  author =        {Shajkofci, Adrian and Liebling, Michael},
  journal =       {IEEE Transactions on Image Processing},
  pages =         {5848--5861},
  publisher =     {IEEE},
  title =         {{Spatially-Variant CNN-Based Point Spread Function
                   Estimation for Blind Deconvolution and Depth
                   Estimation in Optical Microscopy}},
  volume =        {29},
  year =          {2020},
}

@inproceedings{kohler2012recording,
  title={Recording and playback of camera shake: Benchmarking blind deconvolution with a real-world database},
  author={K{\"o}hler, Rolf and Hirsch, Michael and Mohler, Betty and Sch{\"o}lkopf, Bernhard and Harmeling, Stefan},
  booktitle={Computer Vision--ECCV 2012: 12th European Conference on Computer Vision, Florence, Italy, October 7-13, 2012, Proceedings, Part VII 12},
  pages={27--40},
  year={2012},
  organization={Springer}
}

@article{oliveira2013parametric,
  title={Parametric blur estimation for blind restoration of natural images: Linear motion and out-of-focus},
  author={Oliveira, Joao P and Figueiredo, Mario AT and Bioucas-Dias, Jose M},
  journal={IEEE Transactions on Image Processing},
  volume={23},
  number={1},
  pages={466--477},
  year={2013},
  publisher={IEEE}
}

@inproceedings{benichoux2013fundamental,
  title={A fundamental pitfall in blind deconvolution with sparse and shift-invariant priors},
  author={Benichoux, Alexis and Vincent, Emmanuel and Gribonval, R{\'e}mi},
  booktitle={2013 IEEE International Conference on Acoustics, Speech and Signal Processing},
  pages={6108--6112},
  year={2013},
  organization={IEEE}
}

@inproceedings{rombach2022high,
  title={High-resolution image synthesis with latent diffusion models},
  author={Rombach, Robin and Blattmann, Andreas and Lorenz, Dominik and Esser, Patrick and Ommer, Bj{\"o}rn},
  booktitle={Proceedings of the IEEE/CVF conference on computer vision and pattern recognition},
  pages={10684--10695},
  year={2022}
}

@article{efron2011tweedie,
  title={Tweedie’s formula and selection bias},
  author={Efron, Bradley},
  journal={Journal of the American Statistical Association},
  volume={106},
  number={496},
  pages={1602--1614},
  year={2011},
  publisher={Taylor \& Francis}
}

@article{vincent2011connection,
  title={A connection between score matching and denoising autoencoders},
  author={Vincent, Pascal},
  journal={Neural computation},
  volume={23},
  number={7},
  pages={1661--1674},
  year={2011},
  publisher={MIT Press}
}

@Inbook{Skilling1989,
author="Skilling, John",
editor="Skilling, J.",
title="The Eigenvalues of Mega-dimensional Matrices",
bookTitle="Maximum Entropy and Bayesian Methods: Cambridge, England, 1988",
year="1989",
publisher="Springer Netherlands",
address="Dordrecht",
pages="455--466",
isbn="978-94-015-7860-8",
doi="10.1007/978-94-015-7860-8_48",
url="https://doi.org/10.1007/978-94-015-7860-8_48"
}

@article{chen2018neural,
  title={Neural ordinary differential equations},
  author={Chen, Ricky TQ and Rubanova, Yulia and Bettencourt, Jesse and Duvenaud, David K},
  journal={Advances in neural information processing systems},
  volume={31},
  year={2018}
}

@article{debarnot2022deep,
  title={Deep-blur: Blind identification and deblurring with convolutional neural networks},
  author={Debarnot, Valentin and Weiss, Pierre},
  journal={Biological Imaging},
  year={2024}
}

@article{noll1976zernike,
  author =        {Noll, Robert J},
  journal =       {JOSA},
  number =        {3},
  pages =         {207--211},
  publisher =     {Optical Society of America},
  title={         {Zernike polynomials and atmospheric turbulence}},
  volume =        {66},
  year =          {1976},
}

@article{tachella2025deepinverse,
  title = {DeepInverse: A Python package for solving imaging inverse problems with deep learning},
  volume = {10},
  ISSN = {2475-9066},
  url = {http://dx.doi.org/10.21105/joss.08923},
  DOI = {10.21105/joss.08923},
  number = {115},
  journal = {Journal of Open Source Software},
  publisher = {The Open Journal},
  author = {Tachella,  Julián and Terris,  Matthieu and Hurault,  Samuel and Wang,  Andrew and Davy,  Leo and Scanvic,  Jérémy and Sechaud,  Victor and Vo,  Romain and Moreau,  Thomas and Davies,  Thomas and Chen,  Dongdong and Laurent,  Nils and Monroy,  Brayan and Dong,  Jonathan and Hu,  Zhiyuan and Nguyen,  Minh-Hai and Sarron,  Florian and Weiss,  Pierre and Escande,  Paul and Massias,  Mathurin and Modrzyk,  Thibaut and Levac,  Brett and Liaudat,  Tobías I. and Song,  Maxime and Hertrich,  Johannes and Neumayer,  Sebastian and Schramm,  Georg},
  year = {2025},
  month = nov,
  pages = {8923}
}

@article{zhang2021plug,
  title={Plug-and-Play Image Restoration with Deep Denoiser Prior},
  author={Zhang, Kai and Li, Yawei and Zuo, Wangmeng and Zhang, Lei and Van Gool, Luc and Timofte, Radu},
  journal={IEEE Transactions on Pattern Analysis and Machine Intelligence},
  volume={44},
  number={10},
  pages={6360-6376},
  year={2021}
}

@unpublished{nguyen:pnp_unrolled,
  TITLE = {{Comparing Plug-and-Play and Unrolled networks}},
  AUTHOR = {Nguyen, Minh Hai and Weiss, Pierre},
  URL = {https://hal.science/hal-04703008},
  NOTE = {working paper or preprint},
  YEAR = {2024},
  MONTH = Sep,
  HAL_ID = {hal-04703008},
  HAL_VERSION = {v1},
}

@inproceedings{stanczuk2024diffusion,
  title={Diffusion models encode the intrinsic dimension of data manifolds},
  author={Stanczuk, Jan Pawel and Batzolis, Georgios and Deveney, Teo and Sch{\"o}nlieb, Carola-Bibiane},
  booktitle={Forty-first International Conference on Machine Learning},
  year={2024}
}

@inproceedings{karczewski2025diffusion,
title={Diffusion Models as Cartoonists: The Curious Case of High Density Regions},
author={Rafal Karczewski and Markus Heinonen and Vikas Garg},
booktitle={The Thirteenth International Conference on Learning Representations},
year={2025},
url={https://openreview.net/forum?id=RiS2cxpENN}
}

@article{krahmer2021convex,
  title={On the convex geometry of blind deconvolution and matrix completion},
  author={Krahmer, Felix and St{\"o}ger, Dominik},
  journal={Communications on Pure and Applied Mathematics},
  volume={74},
  number={4},
  pages={790--832},
  year={2021},
  publisher={Wiley Online Library}
}

@article{debarnot2023blind,
  title={Blind inverse problems with isolated spikes},
  author={Debarnot, Valentin and Weiss, Pierre},
  journal={Information and Inference: A Journal of the IMA},
  volume={12},
  number={1},
  pages={26--71},
  year={2023},
  publisher={Oxford University Press}
}

@article{ling2018self,
  title={Self-calibration and bilinear inverse problems via linear least squares},
  author={Ling, Shuyang and Strohmer, Thomas},
  journal={SIAM Journal on Imaging Sciences},
  volume={11},
  number={1},
  pages={252--292},
  year={2018},
  publisher={SIAM}
}

@article{candes2005decoding,
  title={Decoding by linear programming},
  author={Candes, Emmanuel J and Tao, Terence},
  journal={IEEE transactions on information theory},
  volume={51},
  number={12},
  pages={4203--4215},
  year={2005},
  publisher={IEEE}
}

@article{vaiter2017model,
  title={Model consistency of partly smooth regularizers},
  author={Vaiter, Samuel and Peyr{\'e}, Gabriel and Fadili, Jalal},
  journal={IEEE Transactions on Information Theory},
  volume={64},
  number={3},
  pages={1725--1737},
  year={2017},
  publisher={IEEE}
}

@InProceedings{blindDPS,
    author    = {Chung, Hyungjin and Kim, Jeongsol and Kim, Sehui and Ye, Jong Chul},
    title     = {Parallel Diffusion Models of Operator and Image for Blind Inverse Problems},
    booktitle = {Proceedings of the IEEE/CVF Conference on Computer Vision and Pattern Recognition (CVPR)},
    month     = {June},
    year      = {2023},
    pages     = {6059-6069}
}

@article{negahban2009unified,
  title={A unified framework for high-dimensional analysis of $ m $-estimators with decomposable regularizers},
  author={Negahban, Sahand and Yu, Bin and Wainwright, Martin J and Ravikumar, Pradeep},
  journal={Advances in neural information processing systems},
  volume={22},
  year={2009}
}

@article{marz2023sampling,
  title={Sampling Rates for {$\ell^1$}-Synthesis},
  author={M{\"a}rz, Maximilian and Boyer, Claire and Kahn, Jonas and Weiss, Pierre},
  journal={Foundations of Computational Mathematics},
  volume={23},
  number={6},
  pages={2089--2150},
  year={2023},
  publisher={Springer}
}

@article{lewis2002active,
  title={Active sets, nonsmoothness, and sensitivity},
  author={Lewis, Adrian S},
  journal={SIAM Journal on Optimization},
  volume={13},
  number={3},
  pages={702--725},
  year={2002},
  publisher={SIAM}
}

@article{liang2014local,
  title={Local linear convergence of forward--backward under partial smoothness},
  author={Liang, Jingwei and Fadili, Jalal M and Peyr{\'e}, Gabriel},
  journal={Advances in neural information processing systems},
  volume={27},
  year={2014}
}

\end{document}